\documentclass[10pt,twocolumn,letterpaper]{article}

\usepackage{cvpr}
\usepackage{times}
\usepackage{url}
\usepackage{epsfig}
\usepackage{graphicx}
\usepackage{amsmath}
\usepackage{amssymb}
\usepackage{float}
\usepackage{epstopdf}
\usepackage{multirow}
\usepackage{subfigure}

\graphicspath{{figures/}}

\providecommand{\keywords}[1]{\textbf{\textit{Keywords ---}} #1}

\usepackage[pagebackref=true,breaklinks=true,letterpaper=true,colorlinks,bookmarks=false]{hyperref}

\cvprfinalcopy 

\def\cvprPaperID{****} 

\ifcvprfinal\fi

\begin{document}

\title{Feature Agglomeration Networks for Single Stage Face Detection}

\author{Jialiang Zhang$^\dagger$$^\ddagger$\thanks{The first two authors contributed equally to this work.},\quad Xiongwei Wu$^\dagger$\footnotemark[1],\quad Jianke Zhu$^\ddagger$,\quad Steven C.H. Hoi$^\dagger$$^\S$\\
$^\dagger$School of Information Systems, Singapore Management University, Singapore\\
$^\ddagger$College of Computer Science and Technology, Zhejiang University, Hangzhou, China\\
$^\S$DeepIR Inc., Beijing, China\\
{\tt\small \{chhoi,xwwu.2015@phdis\}@smu.edu.sg};{\tt\small\{zjialiang,jkzhu\}@zju.edu.cn}\\
}
\maketitle

\begin{abstract}
Recent years have witnessed promising results of face detection using deep learning. Despite making remarkable progresses, face detection in the wild remains an open research challenge especially when detecting faces at vastly different scales and characteristics. In this paper, we propose a novel simple yet effective framework of ``Feature Agglomeration Networks" (FANet) to build a new single stage face detector, which not only achieves state-of-the-art performance but also runs efficiently. As inspired by Feature Pyramid Networks (FPN) \cite{lin2016fpn}, the key idea of our framework is to exploit inherent multi-scale features of a single convolutional neural network by aggregating higher-level semantic feature maps of different scales as contextual cues to augment lower-level feature maps via a hierarchical agglomeration manner at marginal extra computation cost. We further propose a {\it Hierarchical Loss} to effectively train the FANet model. We evaluate the proposed FANet detector on several public face detection benchmarks, including PASCAL face, FDDB and WIDER FACE datasets and achieved state-of-the-art results. Our detector can run in real time for VGA-resolution images on GPU.

\end{abstract}

\vspace{-0.2in}
\section{Introduction}\label{section: introduction}
Face detection is generally the first key step towards face related applications, such as face alignment, face recognition and facial expression
analysis, etc. Despite being studied extensively, detecting faces in the wild remains an open research problem due to various challenges with real-world faces.

Early works of face detection mainly focused on crafting effective features manually and then building powerful classifiers~\cite{viola2004robust}, which are often sub-optimal and may not always achieve satisfactory results. Recent years have witnessed the successful applications of deep learning techniques for face detection tasks~\cite{ren2015faster,liu2016ssd}. Despite being extensively studied, it remains an open challenge for building a fast face detector with high accuracy in any real-world scenario.

In general, face detection can be viewed as a special case of generic object detection~\cite{ren2015faster,liu2016ssd}. Many previous state-of-the-art face detectors inherited a lot of successful techniques from generic object detection, especially for the family of region-based CNN (R-CNN) methods. Among the family of R-CNN based face detectors, there are two major categories of detection frameworks: (i) two-stage detectors (a.k.a. ``proposal-based"), such as Fast R-CNN~\cite{girshick2015fast}, Faster R-CNN~\cite{ren2015faster}, etc; and (ii) single-stage detectors (a.k.a. ``proposal-free") , such as Region Proposal Networks (RPN)~\cite{ren2015faster}, Single-Shot Multibox Detector (SSD)~\cite{liu2016ssd}, etc. The single-stage detection framework enjoys higher inference efficiency, and thus has attracted increasing attention recently due to the high demand of real-time face detectors in real applications. 

Despite enjoying computational advantages, single-stage detectors' performance can may dramaticaly when handling small faces. In order to build a robust detector, there exist two major routes for improvement. One way is to train multi-shot single-scale detectors by using the idea of image pyramid to train multiple separate single-scale detectors each for one specific scale (e.g., the HR detector in~\cite{Hu_2017_CVPR}). However, such approach is computationally expensive since it has to pass a very deep network multiple times during inference. Another way is to train a single-shot multi-scale detector by exploiting multi-scale feature representations of a deep convolutional network, requiring only a single pass to the network during inference. For example, S3FD~\cite{Zhang_2017_ICCV} follows the second approach by extending SSD~\cite{liu2016ssd} for face detection.

Though achieving promising performance, S3FD shares the similar drawback of SSD-style detection frameworks, where each of multi-scale feature maps is used alone for prediction and thus a high-resolution semantically weak feature map  may fail to perform accurate predictions. Inspired by the recent success of Feature Pyramid Networks (FPN) \cite{lin2016fpn} for generic object detection, we propose a novel simple yet effective detection framework of ``Feature Agglomeration Networks" (FANet) to overcome this by combining low-resolution semantically strong features with high-resolution semantically weak features. In particular, FANet aims to create a hierarchical feature pyramid with rich semantics at all scales to boost the prediction performance of high-resolution feature maps using rich contextual cues from low-resolution semantically strong features.
Unlike FPN that creates feature pyramid using the skip connection module, we propose a novel ``Agglomeration Connection" module to create a new hierarchical feature pyramid for FANet. Besides, a new Hierarchical Loss (HL) is presented to train the FANet model effectively in an end-to-end manner. We conduct extensive experiments on several public face detection benchmarks to validate the efficacy of our proposed FANet structure as well as the HL training scheme.

As a summary, the main contributions of this paper include the following
\begin{itemize}
	\item We introduce Agglomeration Connection module to enhance the feature representation power in high-resolution shallow layers.
	\item We propose a simple yet effective framework of Feature Agglomeration Networks (FANet) for single stage face detection, which creates a new hierarchical effective feature pyramid with rich semantics at all scales;
	\item An effective Hierarchical Loss based training scheme is presented to train the proposed FANet model in an end-to-end manner, which guids a more stable and better training for discriminative features;
	\item Comprehensive experiments are carried out on several public Face Detection benchmarks to demonstrate the superiority of the proposed FANet framework.
\end{itemize}

\section{Related Work}
\textbf{Generic Object Detection.}
As a special case of generic object detection, many face detectors inherit successful techniques for generic object detection~\cite{Zhang_2017_ICCV,liu2016ssd,sun2018face}. There are two major categories of Region-based CNN variants for object detection: (i) two-stage detection systems where proposals are generated in the first stage and further classified in the second stage; and (ii) single-stage detection systems where the object detection and classification are performed simultaneously from the feature maps without a separate proposal generation stage. The two-stage detection systems include Fast R-CNN~\cite{girshick2015fast}, Faster R-CNN~\cite{ren2015faster} and their variants, and the single-stage detection systems include YOLO~\cite{redmon2016you}, RPN~\cite{ren2015faster}, SSD~\cite{liu2016ssd}, etc. Our detector essentially belongs to the single-stage detection framework.

\textbf{Multi-shot single-scale Face Detector.} To detect faces with a large range of scales, one way is to train multiple detectors each of which targets for a specific scale. Hu et al. \cite{Hu_2017_CVPR} trained multiple separate RPN detectors for different scales and made inference using image pyramids. However, their method is very time-consuming since images are required to pass a very deep network multiple times during inference. Hao et al. \cite{hao2017scale} learned a Scale Aware Network by estimating face scales in images and built image pyramids according to the estimated values. Although avoiding computation cost to some extent, multiple passes are still required if faces of varied ranges of scales presented in one image. Due to high computational cost, such paradigm is not suitable for real-time applications.

\textbf{Single-shot multi-scale Face Detector.} Single-shot multi-box detector (SSD)~\cite{liu2016ssd} applies multi-scale feature representations for detecting different scales and thus only a single pass is required. S3FD~\cite{Zhang_2017_ICCV} inherits SSD framework with carefully designed scale-aware anchors. However, S3FD shares the same limitation of SSD, where each feature is used alone for prediction and as a consequence, high-resolution features may fail to provide robust prediction due to the weak semantics. As inspired by FPN~\cite{lin2016fpn}, we propose a new framework of FANet to effectively address the limitation of S3FD by aggregating low-resolution semantically strong features with high-resolution semantically weak features using the proposed Agglomeration Connection module. 

\textbf{Context Modeling} Contextual information is important to improve face detection. In \cite{Zagoruyko2016Multipath}, context is modeled by enlarging the window around proposals. For face detection, CMS-RCNN \cite{zhu2017cms} utilizes a larger window with the cost of duplicating the classification head. This increases the memory requirement as well as detection time. SSH \cite{najibi2017ssh} uses the idea of inception module to create context module. While in our Agglomeration Connection module, other than an inception-like context module, we also incorporate the semantics from a deeper feature maps through agglomeration manner. 

\textbf{Feature Pyramid.} Feature pyramid is a structure which combines semantically weak features with semantically strong features using skip-connection. IoN \cite{bell2016inside} extracts RoI features from different feature maps and concatenates them together. HyperNet \cite{kong2016hypernet} makes prediction on a Hyper Feature Map produced by aggregating multi-scale feature maps. DSSD \cite{fu2017dssd} and RON \cite{KongtCVPR2017} also apply the idea of lateral skip connection to create feature pyramids and achieve promising performance. In this paper, we propose a new Agglomerate Connection module which can aggregate multi-scale features more effectively than the skip connection module. Besides, we also introduce a novel Hierarchical Loss on the proposed FANet framework which enables us to train this powerful detector effectively and robustly in an end-to-end approach.

\section{Feature Agglomeration Networks} \label{fan-sec}

In this section, we present the Feature Agglomeration Networks (FANet) framework for face detection. First, we present the overall architecture of FANet. Then we propose the core agglomeration connection module for building FANet. The third part is the detailed configuration of our detector. Finally, the Hierarchical loss will be introduced to guide a more stable and better training in our designed network structure.

\begin{figure*}[!ht]
	\vspace{-0.1in}
	\begin{center}
		\includegraphics[scale=0.53, trim={0cm 0cm 0 2cm}]{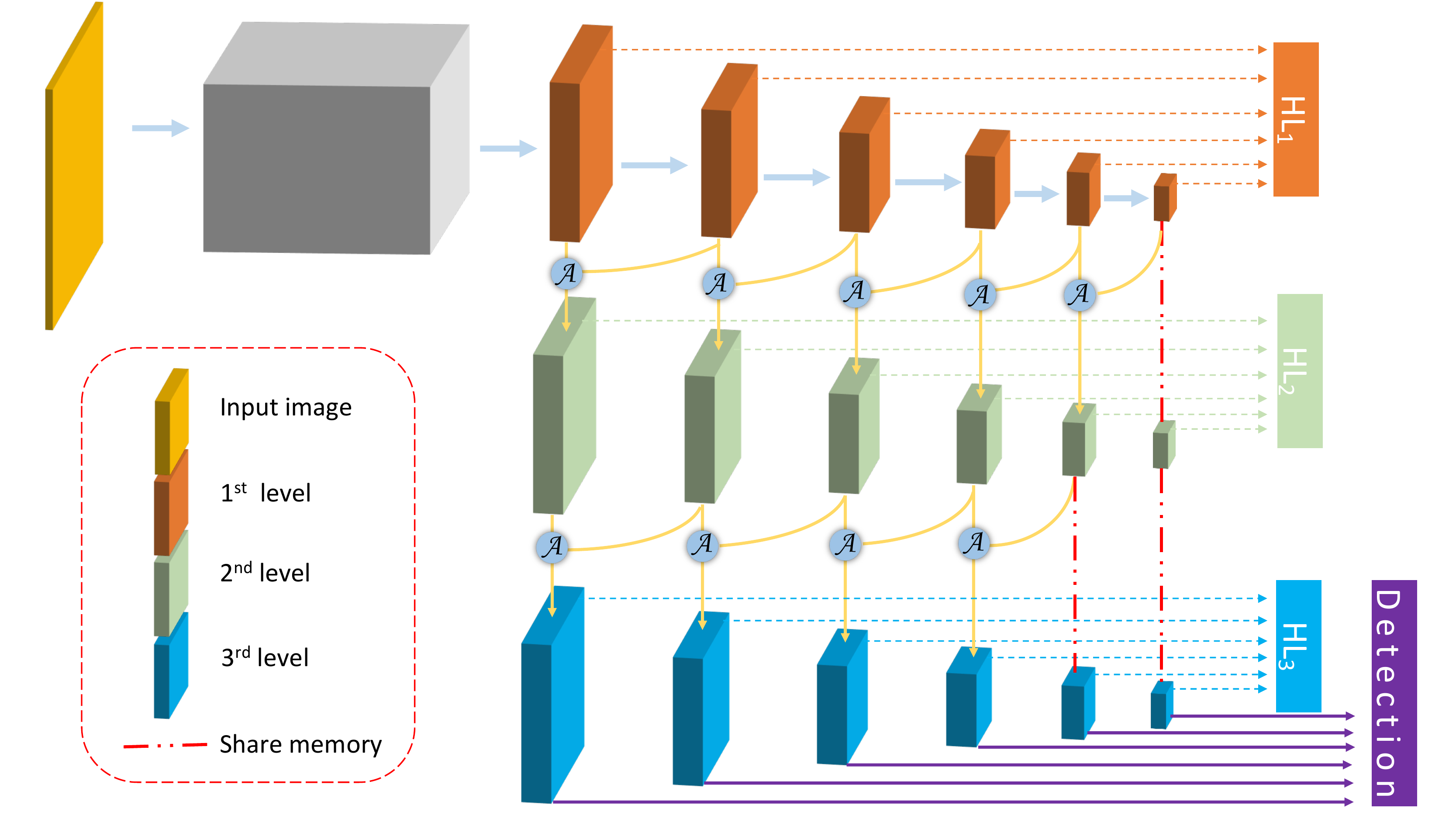} \vspace{-0.1in}
	\end{center}
	\caption{The network architecture of the proposed ``Feature Agglomeration Networks" (FANet). This is a three-level FANet architecture with VGG-16 variant as the backbone CNN network. Hierarchical Loss($HL_i$) accounts for all level feature maps while detection is performed on the last level feature maps.}
	\label{fig:fan structure}
\end{figure*}

\subsection{Overall Architecture} \label{ssec:3.2}

Our goal is to create an effective feature hierarchy with rich semantics at all levels to achieve robust multi-feature detection. Figure. \ref{fig:fan structure} shows our proposed Feature Agglomeration Network (FANet) with 3-level feature hierarchies. The proposed FANet framework is general-purpose. Without loss of generality, in this paper, we consider the widely used VGG16 model as the backbone CNN architecture and SSD as the single stage detector. As shown in Figure. \ref{fig:fan structure}, the detection is performed on $n=6$ layers of feature maps (ranging from index $1$ to $6$). The existing SSD-like detector simply runs detections on the the six feature maps of the first level of feature hierarchy for face detection. By contract, we create the multi-level feature hierarchy with feature agglomeration, and run face detection on the enhanced feature maps (The $3^{rd}$ level, highlighted as blue feature maps in Figure. \ref{fig:fan structure}). Specifically, the proposed Feature Agglomeration operation for an $m$-level FANet ($m\leq n$) can be mathematically defined as follows:
\begin{eqnarray}
\phi_{l}^{1} &=& \mathcal{F}_l(\phi_{l-1}^{1}) \label{eq:1}\\
\phi_{l}^{k} &=& \mathcal{A}_l(\phi_{l}^{k-1},\phi_{l+1}^{k-1}) \quad
k=2,..,m\label{eq:2}
\end{eqnarray}
where $\phi_{l}^{k}$ denotes the feature maps in the $l$-th layer and the $k$-th hierarchy. Specifically, for $k=1$, i.e., the first-level hierarchy, $\phi_{l}^{k}$ is the original feature maps in vanilla S3FD, and $\mathcal{F}_l(\cdot)$ is the non-linear function to transform the feature maps from $l$-th layer to $(l+1)$-th layer, which consists of Convolution, ReLU and Pooling layers, etc. For $k>1$, Eq.(\ref{eq:2}) denotes that $\phi_l^k$ is generated through a feature agglomeration function $\mathcal{A}_l$ to agglomerate two adjacent-layer feature maps in the same hierarchy ($\phi_l^{k-1}$, $\phi_{l+1}^{k-1}$). The agglomeration function $\mathcal{A}_l$ is critical to the performance of the proposed detector. In the following, we propose a novel Agglomeration Connection block for the agglomeration function.

\begin{figure}[htb]
	\vspace{-0.05in}
	\includegraphics[scale=0.25, trim={0cm 2cm 0cm 1cm}]{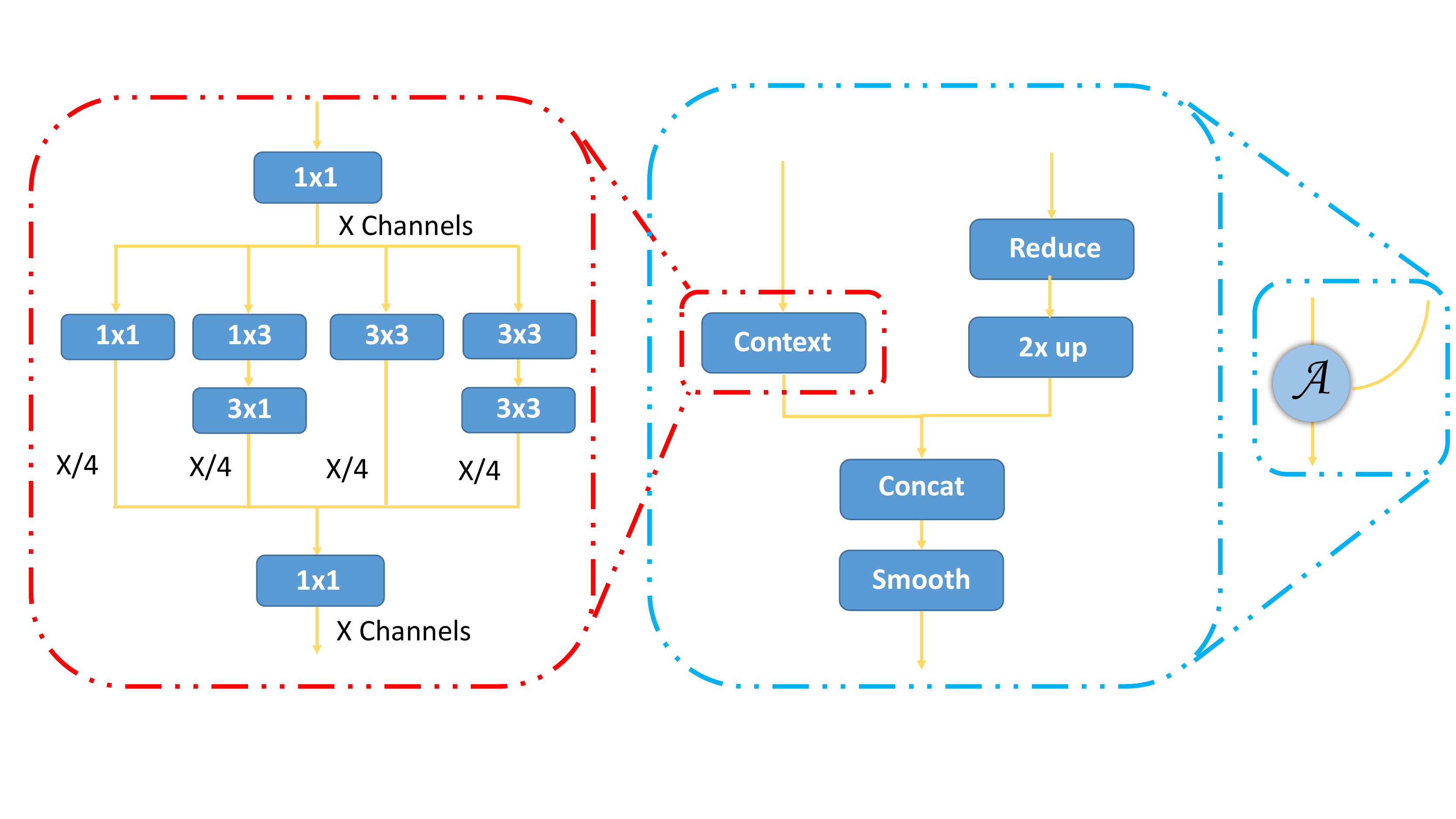}
	\caption{The Agglomeration Connection block ($\mathcal{A}$-block), where the left diagram is a context-aware extraction module.}
	\label{fig:A-block}
\end{figure}

\subsection{Agglomeration Connection}
Figure. \ref{fig:A-block} illustrates the idea of the proposed Agglomeration Connection building block, called ``$\mathcal{A}$-block" for short. It consists of two input feature maps, a shallower feature map $\phi_1$ and a deeper one $\phi_2$. First of all, we notice that feature maps in the shallower layers generally lack of semantics. We thus apply an inception-like module on the shallower feature $\phi_1$ to enhance its feature representation and at the meantime change the output channel to a fixed number $X$ (e.g., $256$). Specifically, as shown in the left diagram of Figure. \ref{fig:A-block}, we use 4 branches with 4 kinds of filters in the shallow feature enhancement block, e.g., $1\times1$, $1\times3$ together with $3\times1$ and $3\times3$. The proportion (output channels) of each branch is $1:1:1:1$, respectively. To ensure the high efficiency of this module, we first apply a $1\times1$ convolution layer to reduce the dimension to $X=256$ channels, and thus our shallow feature enhancement module uses fewer parameters compared with those directly using $3\times3$ filters.

For the deeper feature map $\phi_2$, we first reduce the dimensionality using a $1\times1$ convolution layer by reducing the channel size to $\frac{1}{8}$ of $N$ (e.g., $32$), and then we apply a $2\times$ bilinear upsampling in order to match the same size as $\phi_1$. The final feature of the Agglomeration Connection block is obtained by concating these two features followed with a $3\times3$ convolution smooth layer.

\subsection{Final Detector with Detailed Configurations} \label{ssec:3.3}

The final detection exploits the $m$-th (m=3) hierarchy of feature maps, including a total of six detection layers \{$\mathrm{conv3}\_3^{(3)}$, $\mathrm{conv4}\_3^{(3)}$, $\mathrm{conv5}\_3^{(3)}$,  $\mathrm{conv}\_\mathrm{fc7}^{(3)}$, $\mathrm{conv6}\_2^{(2)}$, $\mathrm{conv7}\_2^{(1)}$\}. The final detection result can be expressed as follows:
\begin{eqnarray}
\mathcal{R} &=& \mathcal{D}(\phi_1^m,\phi_2^{m},...,\phi_n^{m}) \label{eq:3}
\end{eqnarray}
where $\mathcal{D}$ denotes the final detection process including bounding box regression and class prediction followed by Non-Maximum Suppression to obtain the final detection results. As shown in Figure. \ref{fig:fan structure}, the red dotted line denotes the connected 2 blobs share the memory, eg. $\mathrm{conv7}\_2^{(3)}$ and $\mathrm{conv7}\_2^{(2)}$ are identical to $\mathrm{conv7}\_2^{(1)}$, etc.

We also discuss the details of configuring our proposed $3$-level FANet for single-stage face detector.
\if In practice, assuming that detection is performed on $n$ feature maps of a CNN model (specifically n=6 in VGG-16 as used in our experiment), we can design a FANet structure with $m$-level hierarchies where $m\leq6$.

\{$\mathrm{conv3}\_3^{(3)}$, $\mathrm{conv4}\_3^{(3)}$, $\mathrm{conv5}\_3^{(3)}$,  $\mathrm{conv}\_fc7^{(3)}$,
$\mathrm{conv6}\_2^{(2)}$, $\mathrm{conv7}\_2^{(1)}$\}
\fi
In Figure. \ref{fig:fan structure}, six detection layers have strides of $\{4,8,16,32,64,128\}$, respectively. We follow the settings of \cite{Zhang_2017_ICCV}, each of the six feature maps is associated with a specific scale anchor $\{16,32,64,128,256,512\}$ with aspect ratio $1:1$ to detect corresponding scale's faces. Since the shallow feature with high resolution plays a key role in detecting small faces, while deep feature is already with sufficient semantics, we build our FANet structure starting aggregating the feature maps from the $4$-th layer $\mathrm{conv}\_fc7^{(1)}$ instead of $\mathrm{conv7}\_2^{(1)}$. We found it does not hurt the performance while reducing the model complexity.

For anchor-based detectors, we need to match each anchor as a positive or negative according to the ground truth bounding boxes. We adopt the following matching strategy: (i) for each face, the anchor with the best Jaccard overlap is matched; and (ii) each anchor is matched to the face that has Jaccard overlap larger than $0.35$.

{\bf Remark.}
The insight behind hierarchical agglomeration design is that in vanilla SSD, shallower features which are important to detect small faces are semantically weak in feature representation. The $\mathcal{A}$-block hierarchically aggregates semantics information from deeper layers to form a stronger set of hierarchical multi-feature maps.
The ratio of deeper and shallower feature in one $\mathcal{A}$-block is set to $\frac{1}{8}$, which ensures that the deeper feature generally plays a role of providing extra contextual cues. Besides, we notice that the receptive field largely impacts the performance of detecting small faces. As shown in \cite{Hu_2017_CVPR}, either too large or too small receptive field can hurt the performance. The superiority of our structure is that the $\mathcal{A}$-block only incorporates semantics from one deeper layer, and thus we can easily control the receptive field of each feature map through our hierarchical design. This is in contrast to FPN \cite{lin2016fpn}, where a feature map incorporates information from all the deeper layers.

\subsection{Hierarchical Loss} \label{ssec:3.4}

In order to train the proposed FANet effectively, we propose a new loss function called hierarchical loss defined on the proposed FANet structure. The key idea is to define a loss function that accounts for all hierarchies of feature maps, and meanwhile allows to train the entire network effectively in an end-to-end manner. See Figure. \ref{fig:fan structure} for more details.
To this end, we propose the hierarchical loss as follows
\begin{eqnarray}
\label{eq:4}
\mathrm{HL}(\phi_1^m,...,\phi_{n}^{m}) = \sum\limits_{i=1}^{m}{\omega_{i}HL_{i}} = \sum\limits_{i=1}^{m}{\omega_i L(\phi_1^i,...,\phi_{n}^{i})}
\end{eqnarray}
where $\omega_i$ is a weight parameter for the loss of the $i$-th hierarchy.  $\mathrm{L}(\phi_1^i,...,\phi_{n}^{i})$ accounts for the loss on the $i$-th hierarchy, which is SSD \cite{liu2016ssd} multibox loss.
\begin{eqnarray}
\label{eq:5}
\hspace{-0.05in}\mathrm{L}(\phi_1^i,...,\phi_{n}^{i})\hspace{-0.05in} = \hspace{-0.05in}\frac{1}{N_i}\sum\limits_{j=1}^{N_i}\hspace{-0.05in}{\Big(\lambda L_{cls}(y_j,y_j^*) + L_{loc}(\mathbf{p}_j,\mathbf{p}_j^*)\Big)}
\end{eqnarray}
Using the hierarchical loss, we can train the FANet detector end-to-end. Specifically, during training, all the losses are simultaneously computed, and the gradients are back propagated to each hierarchy of feature maps, respectively.

In contrast to the standard loss, the proposed hierarchical loss enjoys some key advantages. On one hand, hierarchical loss plays a crucial role in training the FANet model robustly and effectively. This is because FANet has more newly added parameters than vanilla SSD for optimization, which is not easy to be directly trained with the existing loss in vanilla SSD training. With multiple hierarchies, hierarchical loss guides a better training process which gradually increases the power of feature maps representation. This allows us to supervise the training process hierarchically to obtain more robust features. On the other hand, compared with the standard single loss, the use of hierarchical loss does not incur extra computation cost during inference after the model has been trained.

\subsection{Other Training Strategies}
\label{ssec:3.5}
In this section, we introduce our data augmentation, hard negative mining and other implementation details.

\textbf{Data augmentation.}
We use similar data augmentation strategies as in SSD\cite{liu2016ssd}, like random flip, color distortion, expansion, etc. Besides, we follow the setting of S3FD\cite{Zhang_2017_ICCV},
instead of directly resizing the whole image to a squared patch (e.g., we use $640\times640$ as input size for training), we first crop a squared patch from original image whose scale ranges from $0.3$ to $1$ of the short size of original image. After random cropping, the final patch is resized to $640\times640$.
\textbf{Hard negative mining.} After anchor matching, most of the anchors are assigned as negatives, which results in a significant imbalance between positive and negative samples. We use an online hard negative mining strategy \cite{liu2016ssd} during training. The ratio of negative and positive anchors is at most $3:1$.

\textbf{Other implementation details.}
We choose $\lambda = 3$ in Eq.(\ref{eq:5}) and $\omega_i$ in Eq.(\ref{eq:4}) as uniform for simplicity. The training starts from fine-tuning VGG16 backbone network using SGD with momentum of 0.9, weight decay of 0.0005, and a total batch size of $12$ on two GPUs. The newly added layers are initialized with ``xavier''. We train our FANet for $120$ epochs and a learning rate of $4 \times 10^{-3}$ for first $80$ epochs and continue training for $20$ epochs with $4 \times 10^{-4}$ and $4 \times 10^{-5}$. Our implementation is based on Pytorch~\cite{paszke2017automatic}, and our source code will be made publicly available.

\section{Experiments} \label{ssec:4} 

In this section, we conduct extensive experiments and ablation studies to evaluate the effectiveness of the proposed FANet framework in two folds. Firstly, we examine the impact of several key components including Agglomeration Connection module, the layer-wise Hierarchical Loss, and other techniques used in our solution. Secondly, we compare the proposed FANet face detector with the state-of-the-art face detectors on popular face detection benchmarks and finally evaluate the inference speed of the proposed face detector.

\subsection{Results on WIDER FACE Datasets}

{\bf Dataset.} We conduct model analysis on the WIDER FACE dataset \cite{yang2016wider}, which has 32,203 images with about 400k faces for a large range of scales. It consists of three subsets: $40\%$ for training, $10\%$ for validation, and $50\%$ for testing. The annotations of training and validation sets are online available. According to the difficulty of detection tasks, it has three splits: Easy, Medium and Hard. The evaluation metric is mean average precision (mAP) with Interception-of-Union (IoU) threshold as 0.5. We train FANet on the training set of WIDER Face, and evaluate it on the validation and testing set. If not specified, the results in Table \ref{tab: compareFPN} and \ref{tab:ablation} are obtained by single scale testing in which the shorter size of image is resized to $640$ while keeping image aspect ratio.

\begin{table}[!htp]
	\centering
	\caption{Evaluation of our FANet for learning discriminative features in contrast to vanilla S3FD and a simple FPN. For fair comparison, we only use 2-level hierarchy without context module.}
	\vspace{0.1in}
	\begin{tabular}{l|c|c|c}
		\hline
		Loss  		& \quad Easy \quad	& Medium	& \quad Hard \quad\\
		\hline
		S3FD \\
		\hline
		vanilla 	& 94.5  & 92.9  & 82.8  \\
		w/ FPN 		& 94.3  & 93.1  & 83.8 \\
		\hline
		FANet \\
		\hline
		w/ 2-level 	& 94.8  & 93.6  & 84.4 \\
		w/ 2-level w/ HL  & 94.8	& 93.6    & 85.3 \\
		\hline
	\end{tabular}
	\label{tab: compareFPN}
\end{table}

{\bf Baseline.} We adopt the closely related detector S3FD \cite{Zhang_2017_ICCV} as the baseline to validate the effectiveness of our technique. S3FD achieved the state-of-the-art results on several well-known face detection benchmarks. It inherited the standard SSD framework with carefully designed scale-aware anchors according to effective receptive fields. We follow the same experimental setup in S3FD.

{\bf Agglomeration Connection.} We validate the contribution of \emph{Agglomeration Connection} module with different hierarchies.

\begin{table*}[htb]
	\centering
	\caption{Evaluation Results on WIDER FACE validation set. Shorter-$768$ indicates the single scale testing that the shorter size of image is resized to $768$ while keeping image aspect ratio. We use this to compare the performance with different input size.}
	\begin{tabular}{l|ccccccc|c}
		&&&&&&&&FANet (ours)\\
		\hline
		1-Level    &${\surd}$&&&&&&&  \\
		2-Level    &&${\surd}$&${\surd}$&&&&&  \\
		3-Level    &&&&${\surd}$&${\surd}$&${\surd}$&${\surd}$&${\surd}$ \\
		Context   	&&&&&&${\surd}$&${\surd}$&${\surd}$ \\
		HL       &&&${\surd}$&&${\surd}$&${\surd}$&${\surd}$&${\surd}$\\
		Shorter-768		&&&&&&&${\surd}$&  \\
		Multi-scale 	&&&&&&&&${\surd}$\\
		\hline
		mAP (Easy)	&94.5 &94.8 &94.8 &94.7 &94.8 &95.0 &94.8 &$\mathbf{95.6}$\\
		mAP (Medium)	&92.9 &93.6 &93.6 &93.3 &93.6 &93.9 &93.6 &$\mathbf{94.7}$\\
		mAP (Hard)	&82.8 &84.4 &85.3 &84.8 &85.7 &86.7 &88.4 &$\mathbf{89.5}$\\
		\hline
	\end{tabular}
	\label{tab:ablation}%
\end{table*}%

First, we validate the efficacy of our hierarchical structure design without HL and inception-like context extraction module. In Table \ref{tab:ablation}, with increasing number of hierarchies, the performance of FANet significantly improves, especially for hard cases which our design mainly aims for. Specifically, the performance gains in hard task are $+1.6\%$, $+2.0\%$  for 2-level and 3-level, respectively. See column $1,2,4$. With the context module, the performance further improved, $+1.0\%$ in hard tasks, shown in column $6,7$. Next we will show that Hierarchical Loss is necessary in guiding a better and more effective training with high hierarchical-level Agglomeration Connection.

{\bf Hierarchical Loss.} In this part, we compared the performance of our FANet optimized with and without Hierarchical Loss. In Table \ref{tab:ablation}, the performance of both 2-level and 3-level FANet with Hierarchical Loss gains significant improvement compared with its single loss setting in hard levels ($+0.9\%$ for 2-level, and $+0.9\%$ for 3-level). Besides, 3-level FANet outperforms 2-level FANet consistently, which indicates high level Agglomeration Connection is crucial to improve detection accuracy with Hierarchical Loss optimization method.

{\bf Robust Feature Learning.}
We build ``S3FD w/ FPN" based on S3FD with skip-connection in top-down structure as FPN\cite{lin2016fpn}. In Table \ref{tab: compareFPN}, compared with S3FD, ``S3FD w/ FPN" gains improvement $+1.0\%$ in hard level, which validates the efficacy of feature pyramid for improving feature representation. Our FANet with 2-level HL outperforms ``S3FD w/ FPN" with large margin, $+1.5\%$ in hard task, which demonstrates the superiority of our agglomeration connection over the skip connection.

{\bf Multi-scale Inference.} Multi-scale testing is a widely used technique in object
detection, which can further boost the detection accuracy especially for small objects.
In Table \ref{tab:ablation}, We show that with larger input size eg. $768$ compared with $640$, it can significantly improve the performance in hard task. The final multi-scale testing result of our FANet is shown in the last column.

\begin{table}[!ht]
	\centering
	\caption{Evaluation on WIDER FACE validation set (mAP).}
	\vspace{0.1in}
	\scalebox{0.8}{
		\begin{tabular}{l|c|c|c|c}
			\hline
			Algorithms    & Backbone & Easy & Med & Hard \\
			\hline
			MTCNN \cite{zhang2016joint} & - &   84.8    &   82.5    &  59.8\\
			LDCF+ \cite{ohn2016boost} & - &   79.0    &   76.9    &  52.2\\
			\hline
			ScaleFace \cite{DBLP:journals/corr/YangXLT17} & ResNet50 &   86.8    &   86.7    &  77.2\\
			HR \cite{Hu_2017_CVPR}&ResNet101 & 92.5 & 91.0 & 80.6\\
			Face R-FCN \cite{DBLP:journals/corr/abs-1709-05256}& ResNet101 & 94.7 & 93.5 & 87.4\\
			Zhu\cite{zhu2018seeing}& ResNet101 & 94.9 & 93.3 & 86.1 \\
			\hline
			CMS-RCNN \cite{zhu2017cms}&  VGG16 &  89.9     &  87.4     &  62.4\\
			MSCNN \cite{cai2016unified}&  VGG16 &  91.6     &  90.3     &  80.2\\
			Face R-CNN \cite{wang2017face}&  VGG19 &  93.7     &  92.1     &  83.1\\
			SSH \cite{najibi2017ssh}& VGG16  &  93.1     &  92.1     &  84.5\\
			S3FD\cite{Zhang_2017_ICCV}& VGG16&  93.7    &  92.5  & 85.9 \\
			FANet(ours)&VGG16 &$\mathbf{95.6}$& $\mathbf{94.7}$ & $\mathbf{89.5}$\\
			\hline
	\end{tabular}}
	\label{tab:overall}
\end{table}

\begin{figure*}[ht]\hspace{-0.2in}
	
	\hspace{-0.2in}
	\subfigure[Easy] {
		\includegraphics[width=0.35\textwidth, trim={0cm 0cm 0  0}]{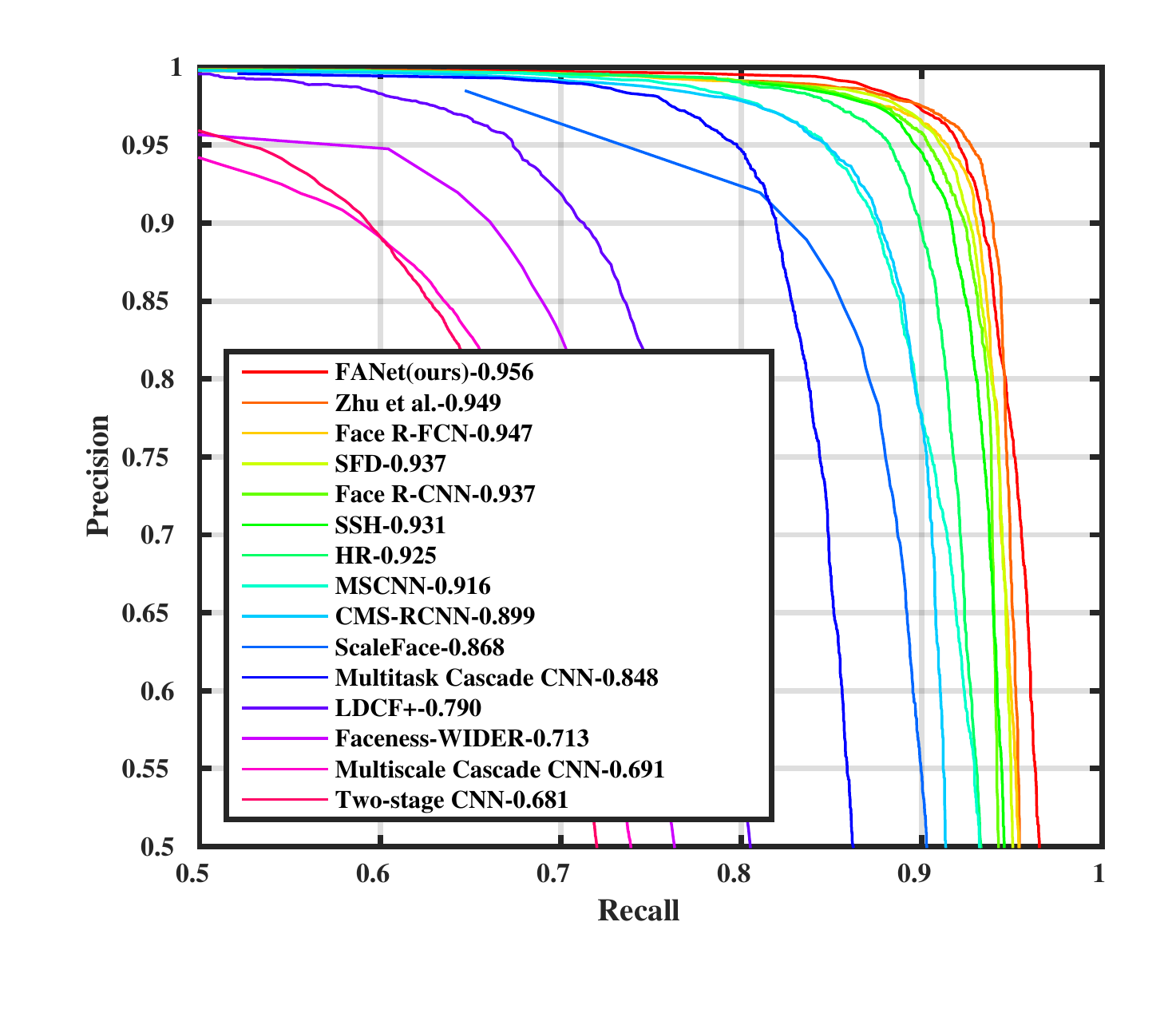}}
	\subfigure[Medium]{
		\includegraphics[width=0.35\textwidth]{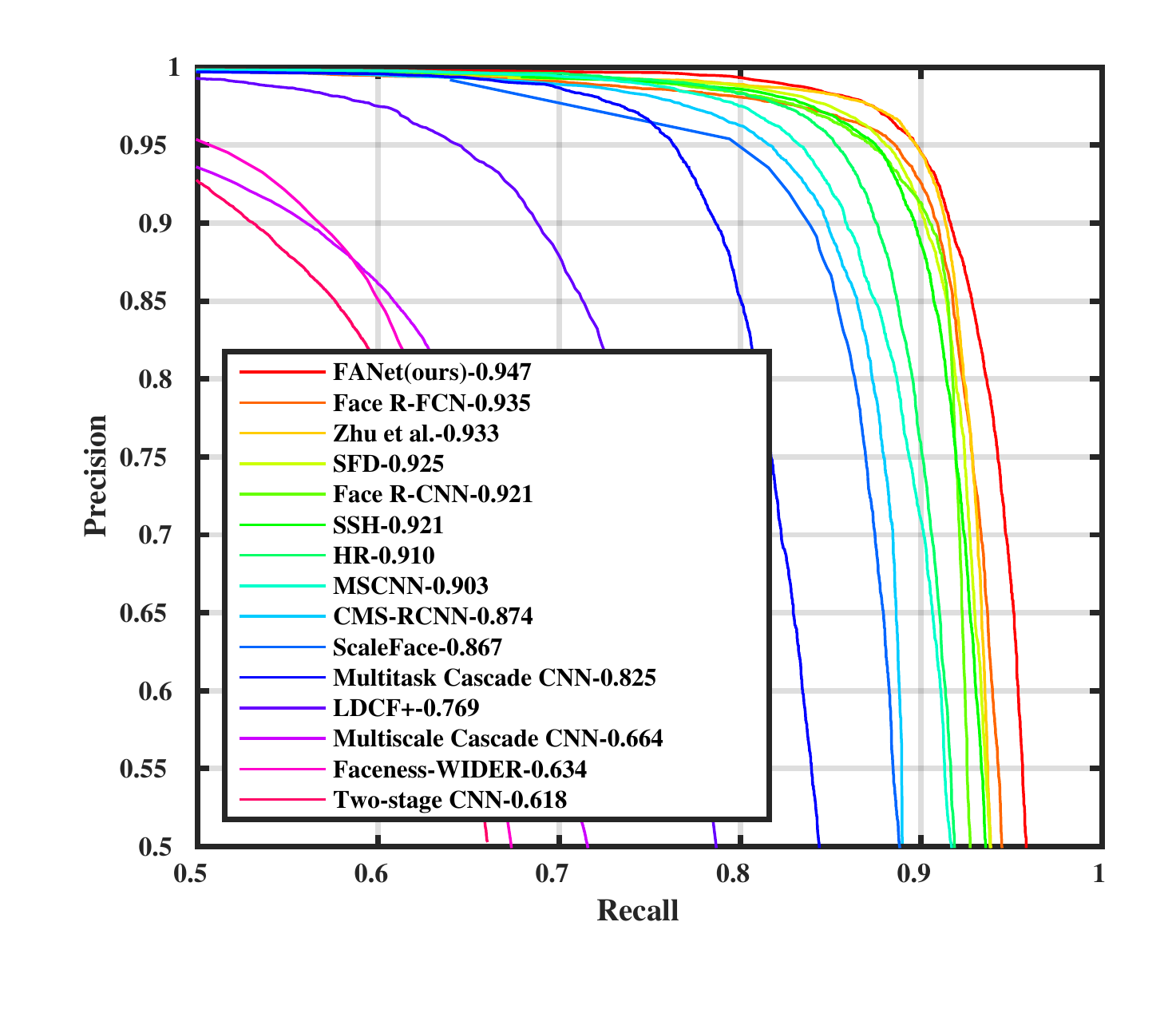}}
	\subfigure[Hard]{
		\includegraphics[width=0.35\textwidth, trim={0cm 0cm 0  0}]{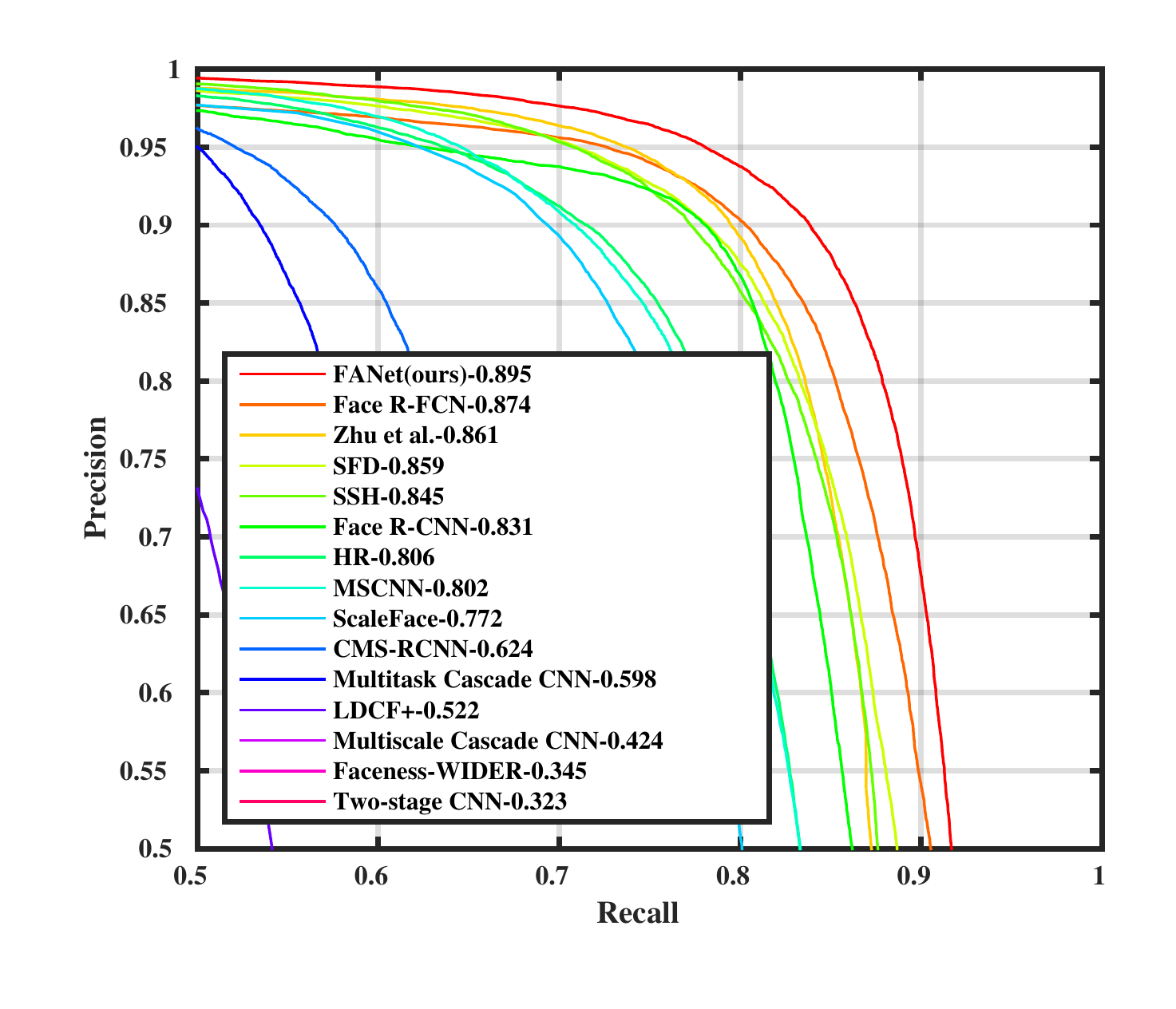}}
	\caption{Evaluation of various state-of-the-art methods on the validation set of WIDER FACE}\label{fig:widerface val}
	
	\subfigure[Easy] {
		\includegraphics[width=0.33\textwidth, trim={1.5cm 0.1cm 0  0}]{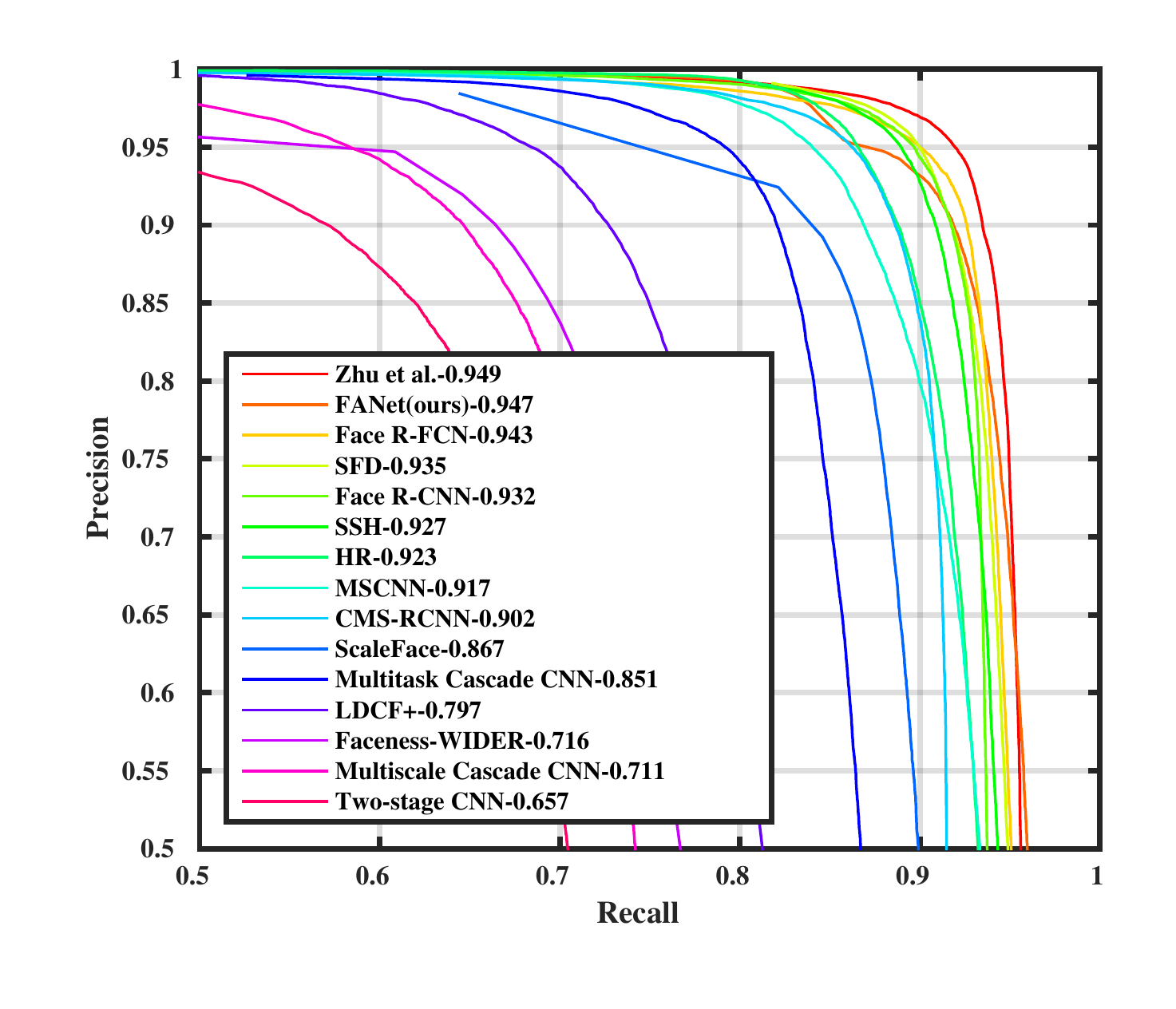}}
	\subfigure[Medium]{
		\includegraphics[width=0.35\textwidth, trim={0.5cm 0 0  0}]{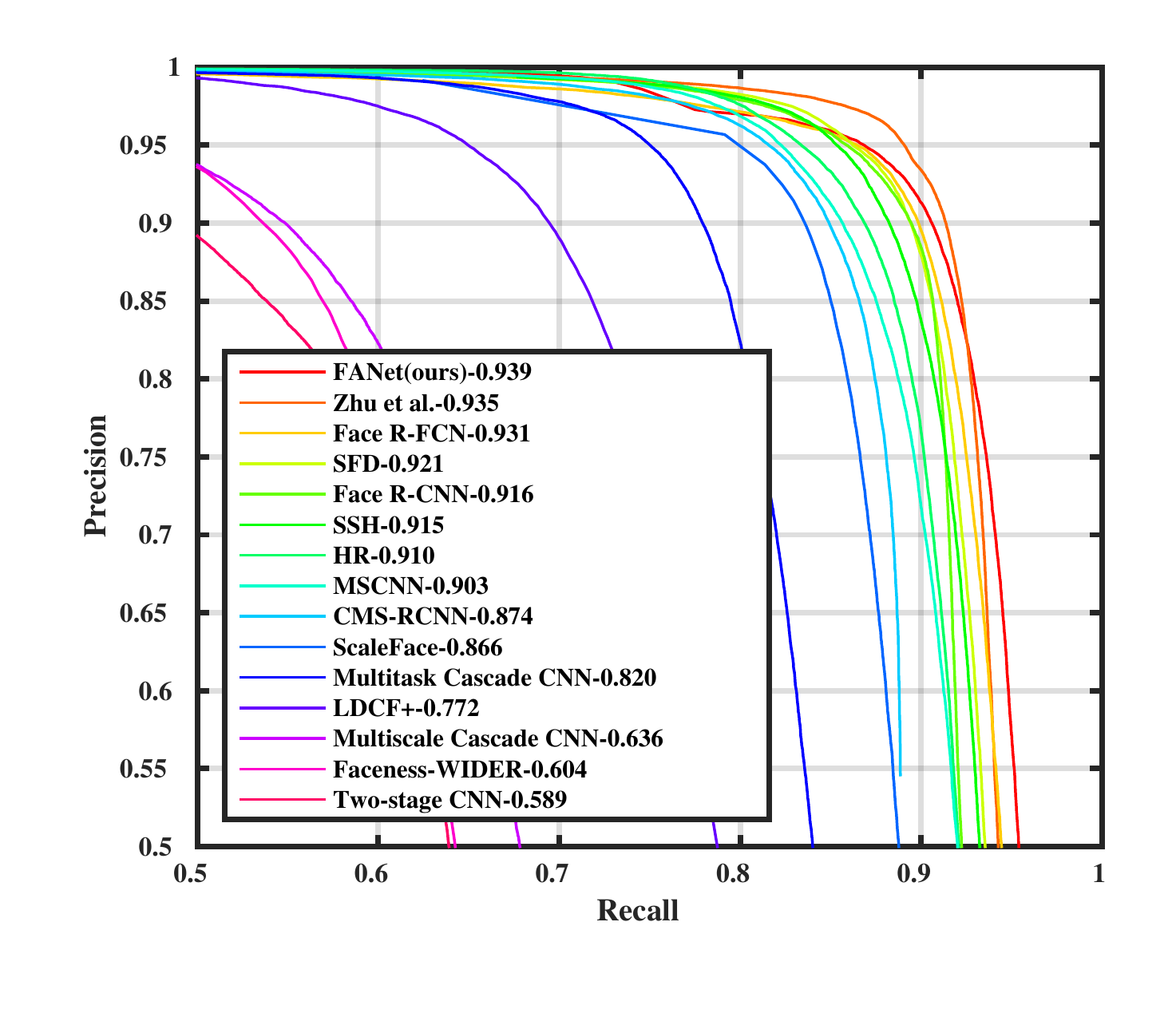}}
	\subfigure[Hard]{
		\includegraphics[width=0.35\textwidth, trim={0.8cm 0 0  0}]{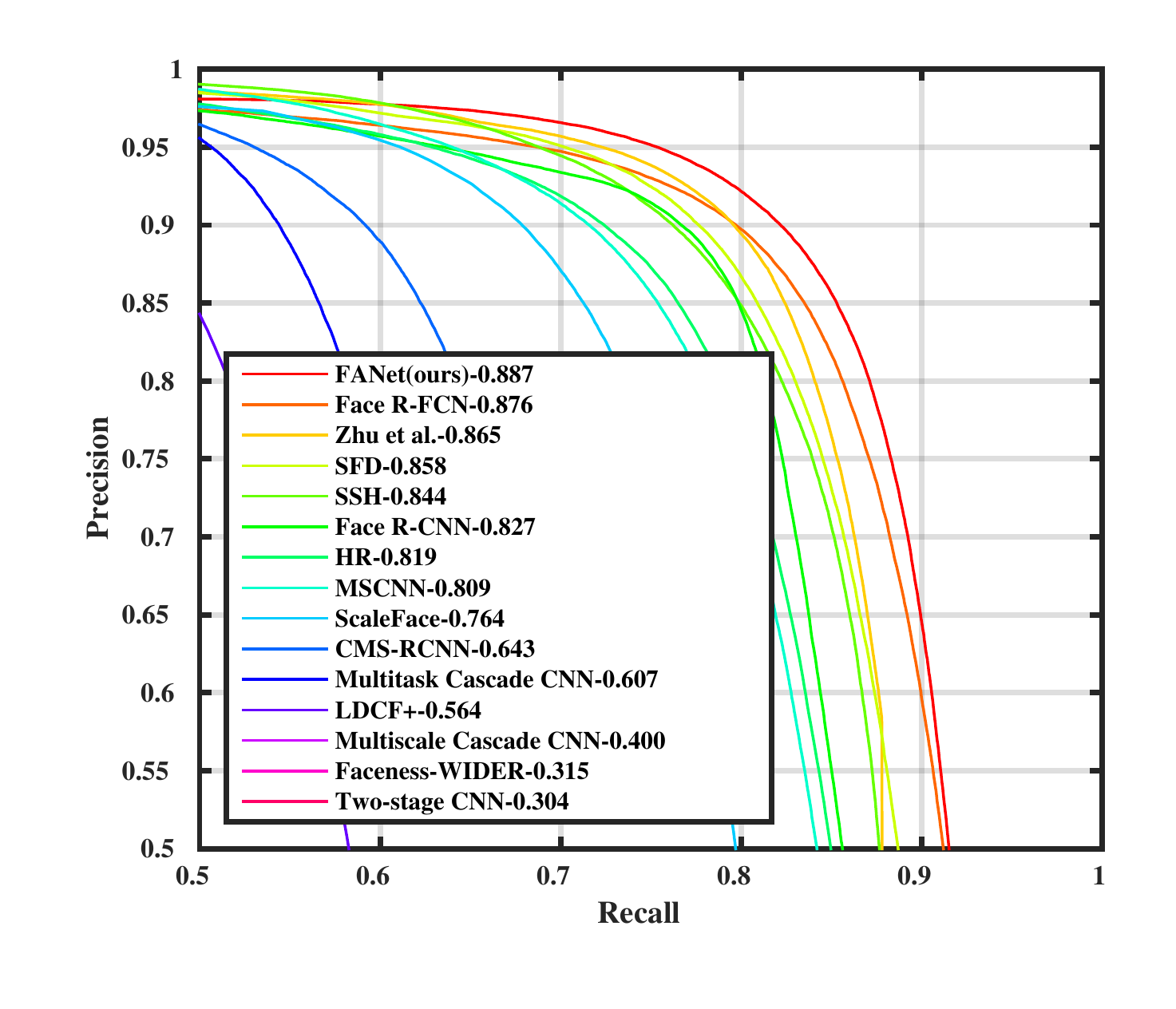}}
	\caption{Evaluation of various state-of-the-art methods on the test set of WIDER FACE}\label{fig:widerface test}
\end{figure*}

{\bf Comparisons with the State of the Art.}\label{sec:wider} Our Final FANet model is trained with 3-level HL. Figure. \ref{fig:widerface val} and Figure. \ref{fig:widerface test} show the precision-recall curves on WIDER Face evaluation and test set and Table \ref{tab:overall} summarizes the state-of-the-art results on the WIDER Face validation and test set.

Our FANet reaches the state-of-the-art result among all detectors.  WIDER FACE is a very challenging face benchmark and the results strongly prove the effectiveness of FANet in handling high scale variances, especially for small faces.

\subsection{Evaluation on Other Public Face Benchmarks.}

In addition to the WIDER FACE data set, we also want to examine the generalization performance of the proposed face detector for other data sets. We thus test the pre-trained FANet face detector on another two popular face detection benchmarks.

\begin{figure*}[!htb]
	\subfigure[FDDB Discrete ROC Curves] {
		\includegraphics[width=0.5\textwidth]{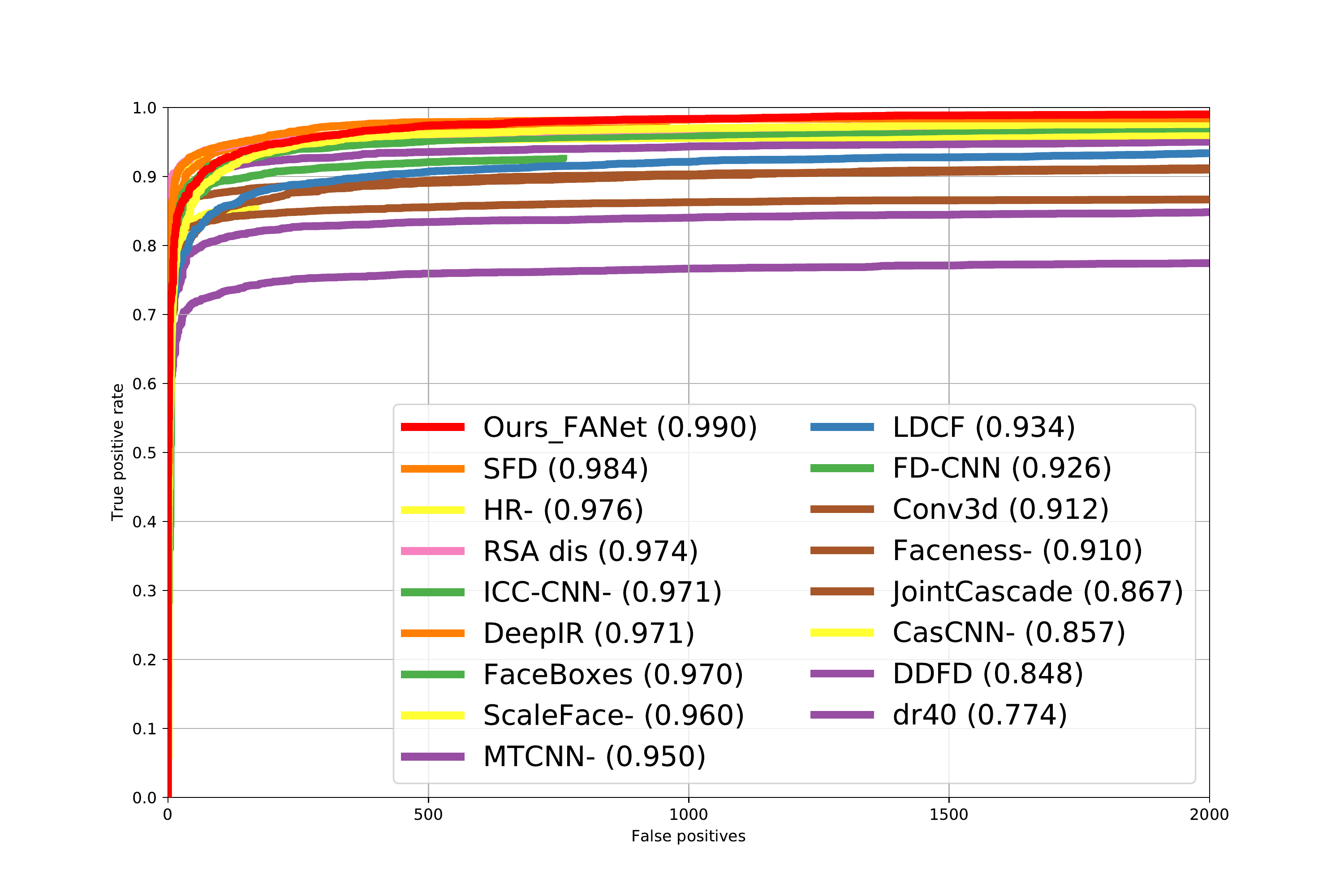}}
	\subfigure[FDDB Continuous ROC Curves]{
		\includegraphics[width=0.5\textwidth]{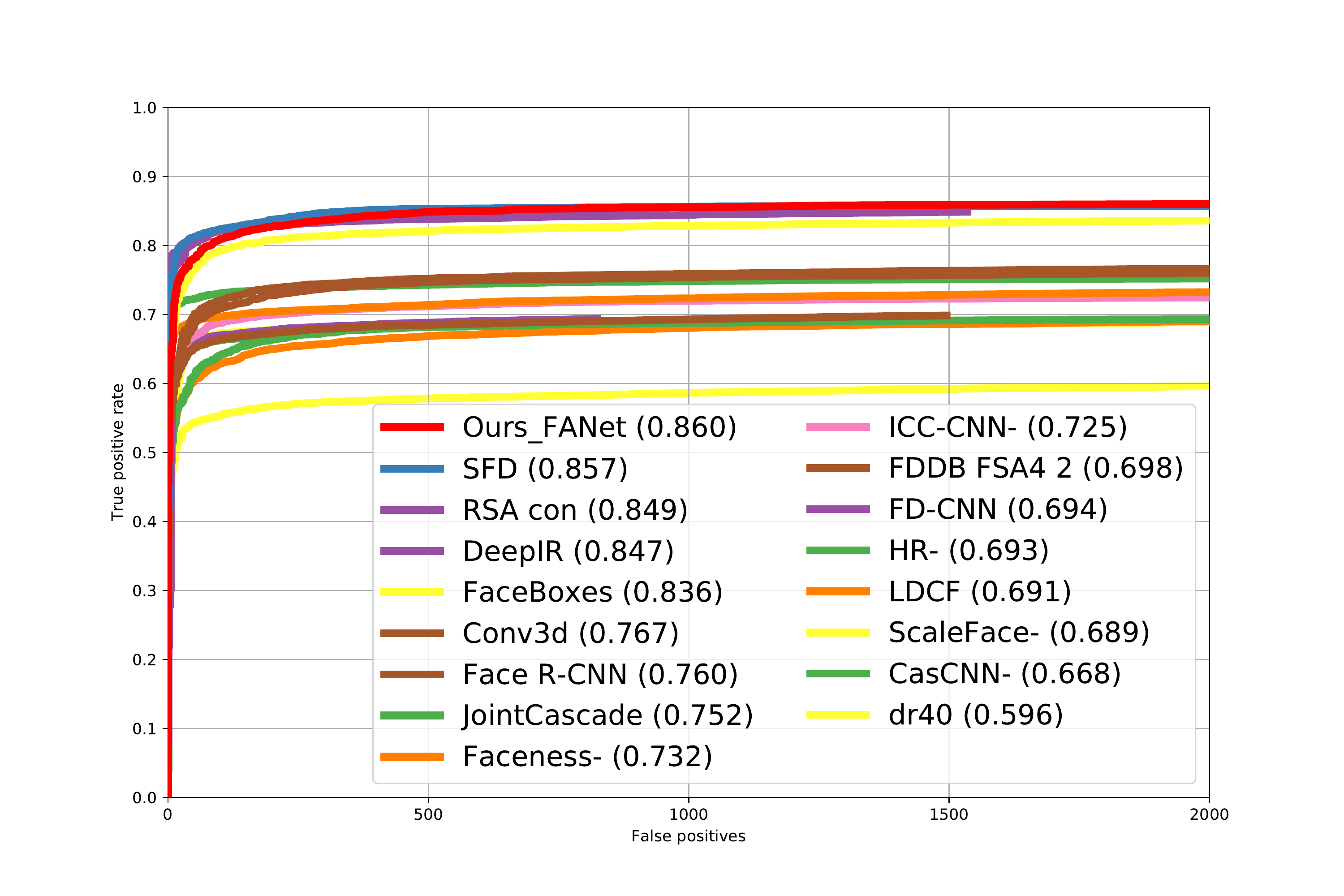}}
	\caption{Evaluation on FDDB face detection benchmarks. Recall@fp2000}\label{fig:fddb}
\end{figure*}

{\bf FDDB.} The Face Detection Data set and Benchmark (FDDB)\cite{fddbTech} is a well-known face detection benchmark with 5,171 faces in 2,845 images. We compare our FANet detector trained on the WIDER FACE training set with other published results on FDDB. Figure. \ref{fig:fddb} shows the evaluation results, in which our FANet detector achieves the state-of-the-art performance on both discrete and continuous ROC curves.

\begin{figure}[ht]
	\begin{center}
		\includegraphics[width=1.1\linewidth]{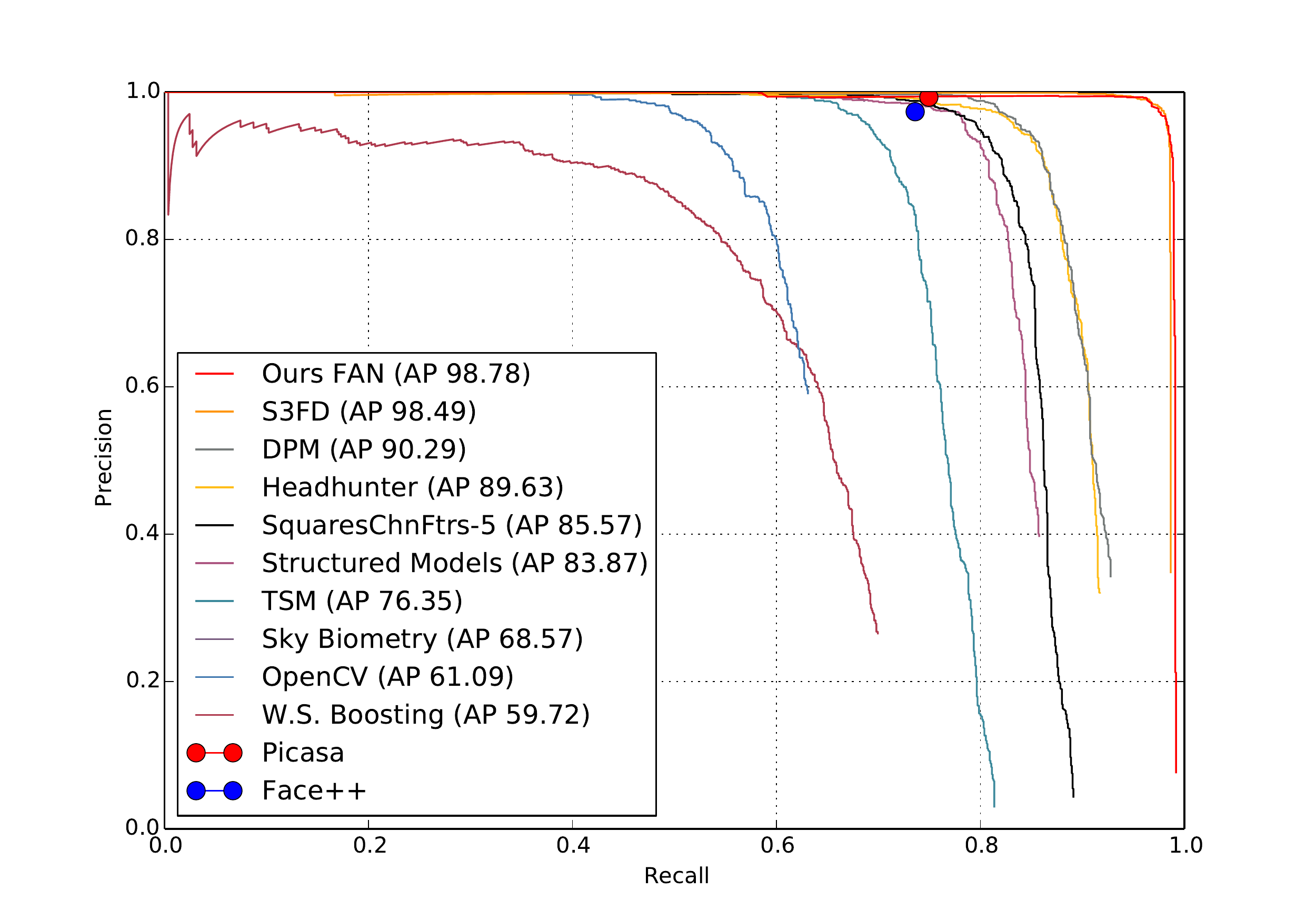}
	\end{center}\vspace{-0.1in}
	\caption{Benchmark evaluation on PASCAL FACE dataset.}\label{fig:pascal}
\end{figure}

{\bf PASCAL FACE.} This dataset was collected from PASCAL person
layout test set \cite{Everingham10}, with 1,335 labeled faces in 851 images. Figure. \ref{fig:pascal} shows the evaluation results of the precision-recall curves. Among all the existing methods, the proposed FANet achieved the best mAP (98.78\%), which outperforms the previous state-of-the-art detectors S3FD (98.45\%), and significantly beats the other submitted methods.
\cite{yang2015facial,chen2016supervised}.

\subsection{Inference Speed}

Our FANet detector is a single-stage detector and thus enjoys high inference speed. It runs in real-time inference speed with 35.6 FPS for VGA-resolution input images on a computing environment with NVIDIA GPU GTX 1080ti and CuDNN-v6.

\section{Conclusion}

This paper proposed a novel framework of ``Feature Agglomeration Networks" (FANet) for building single stage face detectors. The proposed FANet based face detector achieves the state-of-the-art performance on several well-known face detection benchmarks, yet still enjoys real-time inference speed on GPU due to the nature of the single-stage detection framework. FANet introduces two key novel components:
(i) the ``Agglomeration Connection" module for context aware feature enhancing and multi-scale features agglomeration with hierarchical structure, which effectively handles scale variance in face detection; and (ii) the Hierarchical Loss to guide a more stable and better training in an end-to-end manner. We noted that the general idea of the Feature Agglomeration Networks is perhaps not restricted to face detection tasks, and might also be beneficial to other types of object detection tasks. For future work, we will explore the extension of the Feature Agglomeration Networks (FANet) framework for more other object detection tasks in computer vision, including generic object detection or specialized object detection tasks in other domains, such as pedestrian detection, vehicle detection, etc.

\subsection*{Acknowledgements}
The authors would like to acknowledge the assistance and collaboration from colleagues of DeepIR Inc.

\bibliographystyle{ieee}
\bibliography{FANet}

\clearpage

\section*{Appendix}
\begin{figure*}[ht]\hspace{-0.2in}
	
	\vspace{+0.2in}
	\hspace{-0.2in}
	\subfigure[Easy] {\label{sub:easy}
		\includegraphics[width=0.35\textwidth, trim={0.5cm 0 0cm 0}]{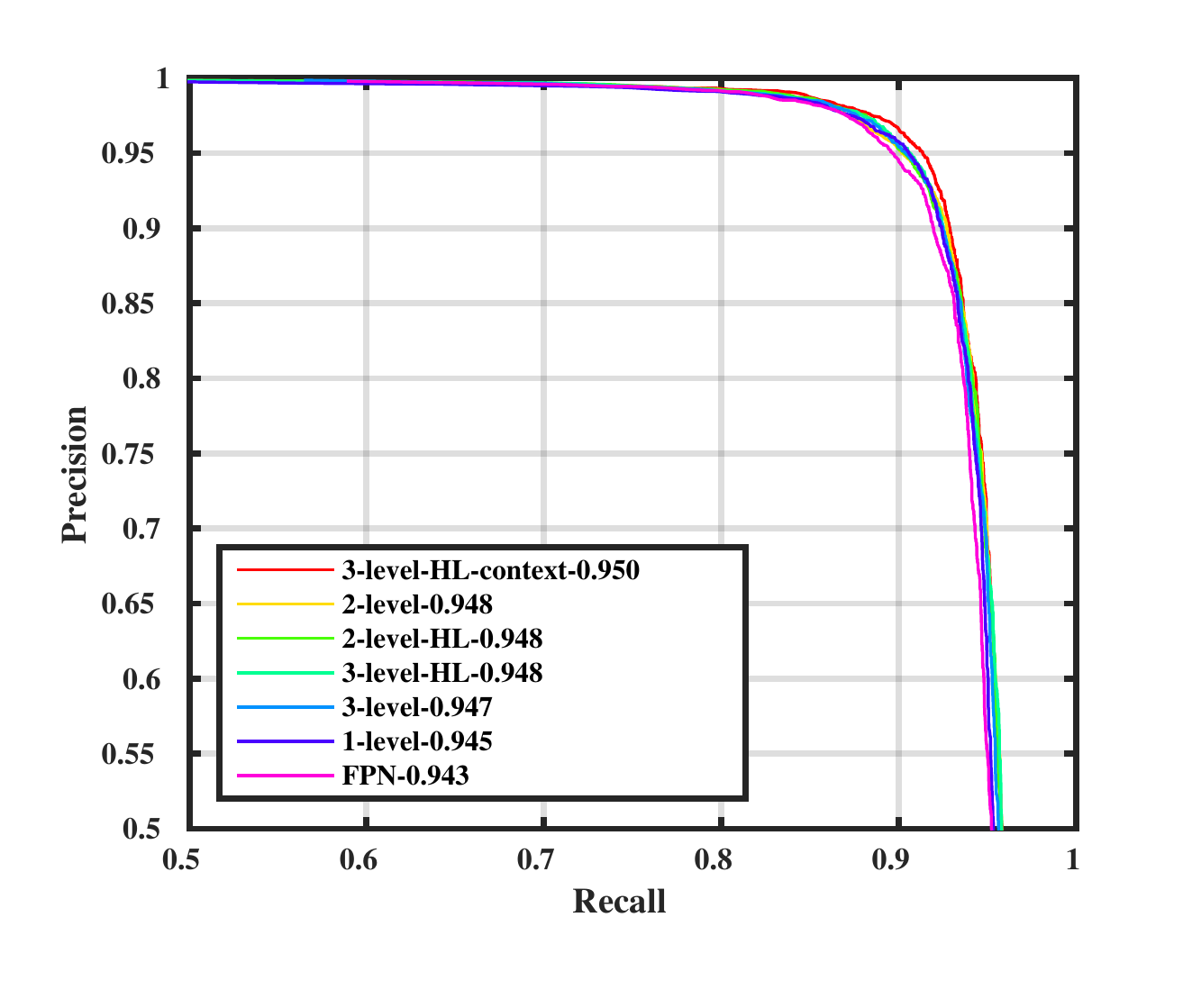}}
	\subfigure[Medium]{\label{sub:medium}
		\includegraphics[width=0.35\textwidth, trim={0.5cm 0 0cm 0}]{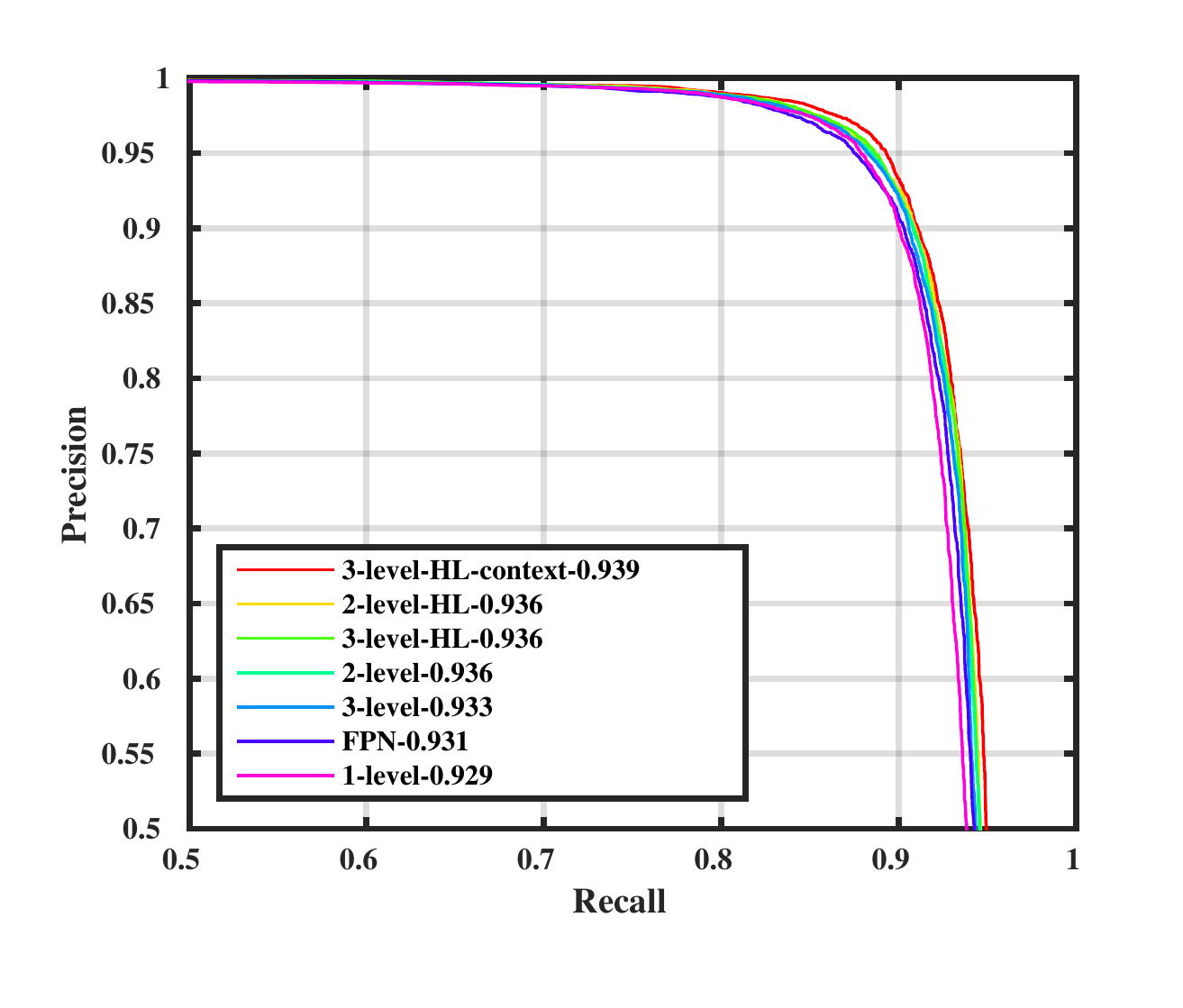}}
	\subfigure[Hard]{\label{sub:hard}
		\includegraphics[width=0.35\textwidth, trim={0.5cm 0 0cm 0}]{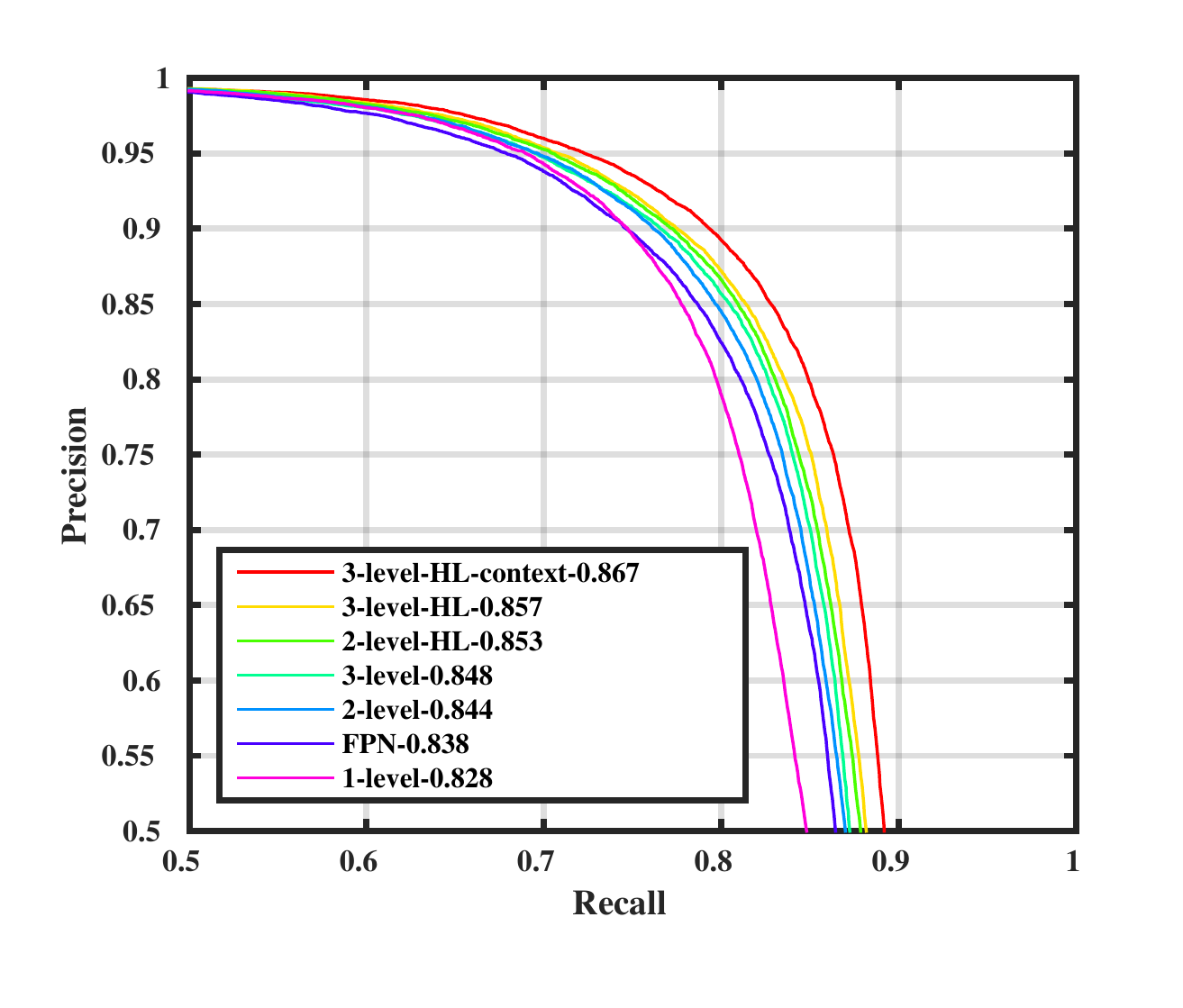}}
	\caption{Precision-recall curves of ablation experiments on the validation set of WIDER Face}\label{fig:ablation} 
	
\end{figure*}

We first showed the precision-recall curves of our ablation experiments, Figure. \ref{fig:ablation}. Then the Speed Accuracy tradeoff and the Model-Size Accuracy tradeoff of our FANet are analyzed. Finally, we demonstrate some qualitative results of our detector. In Figure. \ref{fig:tradeoff}, the result is obtained on the hard task of WIDER face \cite{yang2016wider} evaluation set \ref{sub:hard} by single scale testing in which the shorter size of image is resized to $640$ while keeping image aspect ratio. We test the result under a computing environment with NVIDIA GPU GTX 1080ti and CuDNN-v6.

\section{Precision-recall curves}
Figure. \ref{fig:ablation} showed the precision-recall curves of our ablation experiments.

\section{Speed Accuracy Tradeoff}
In this section, we compared the Speed Accuracy tradeoff between vanilla S3FD \cite{Zhang_2017_ICCV} structure, FPN \cite{lin2016fpn} and our FANet. As shown in Figure. \ref{speed}, our 2-level is a good tradeoff between speed and accuracy. Specifically, it achieves comparable speed as FPN structure (45.4 fps vs 45.9 fps), while getting much better result than FPN (85.3 vs 83.8). Our final FANet improved +3.9\% over vanilla S3FD while still reaching the real-time speed.

\section{Model-Size Accuracy Tradeoff}
\begin{figure}[t]
	\subfigure[Speed accuracy tradeoff.] {\label{speed}
		\includegraphics[width=0.45\textwidth, trim={1cm 0cm 0.5cm +6cm}]{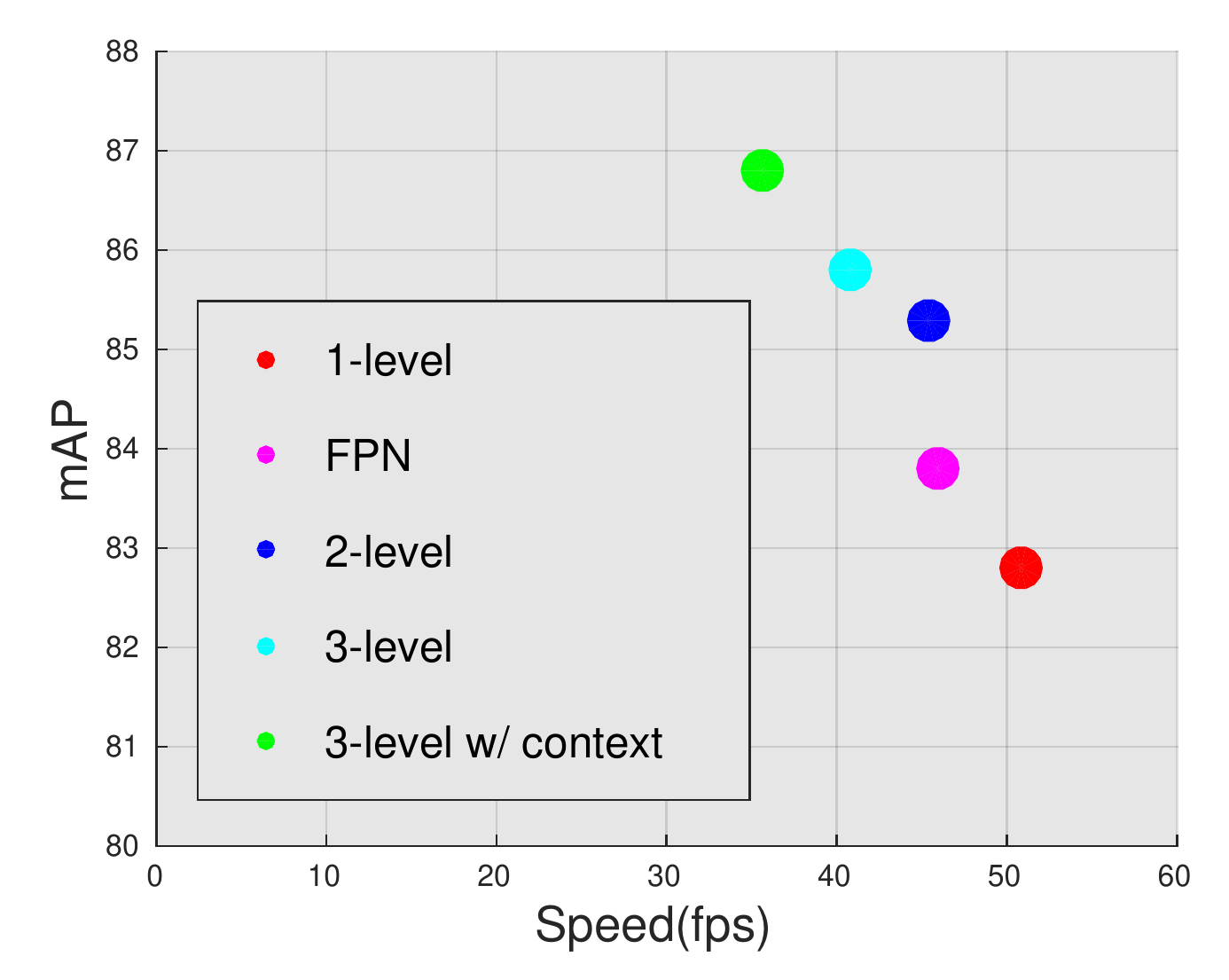}}
	\subfigure[ModelSize accuracy tradeoff.]{\label{modelsize}
		\includegraphics[width=0.45\textwidth, trim={1cm 0cm 0.5cm 0}]{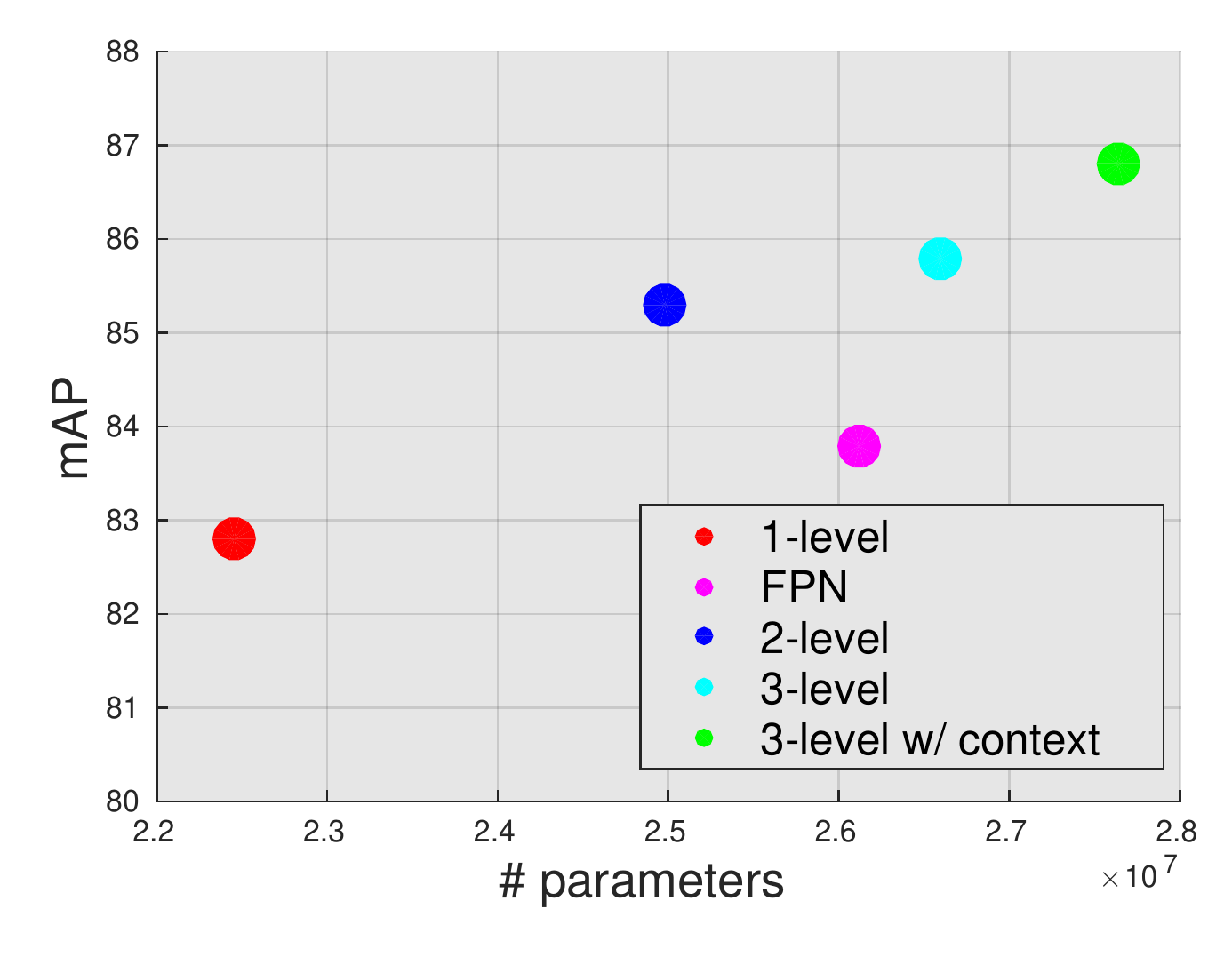}}
	\caption{Model performance analysis. 1-level is of the same structure as S3FD and 3-level w/ context is our final FANet.} \label{fig:tradeoff}
\end{figure}

In this section, we showed the Model Size Accuracy tradeoff of our FANet. As shown in Figure. \ref{modelsize}, our 2-level structure achieves better results than FPN while still having less parameters than it. Compared with vanilla S3FD, our 2-level has only 11\% more parameters while reaching much better results (85.3 vs 82.8). The final FANet gets significant improved results +3.9\% with tolerable increased parameters.

\section{Qualitative Analysis}
\begin{figure*}[!ht]
	\begin{center}
		\includegraphics[scale=0.57, trim={5.5cm 0 5cm 0cm}]{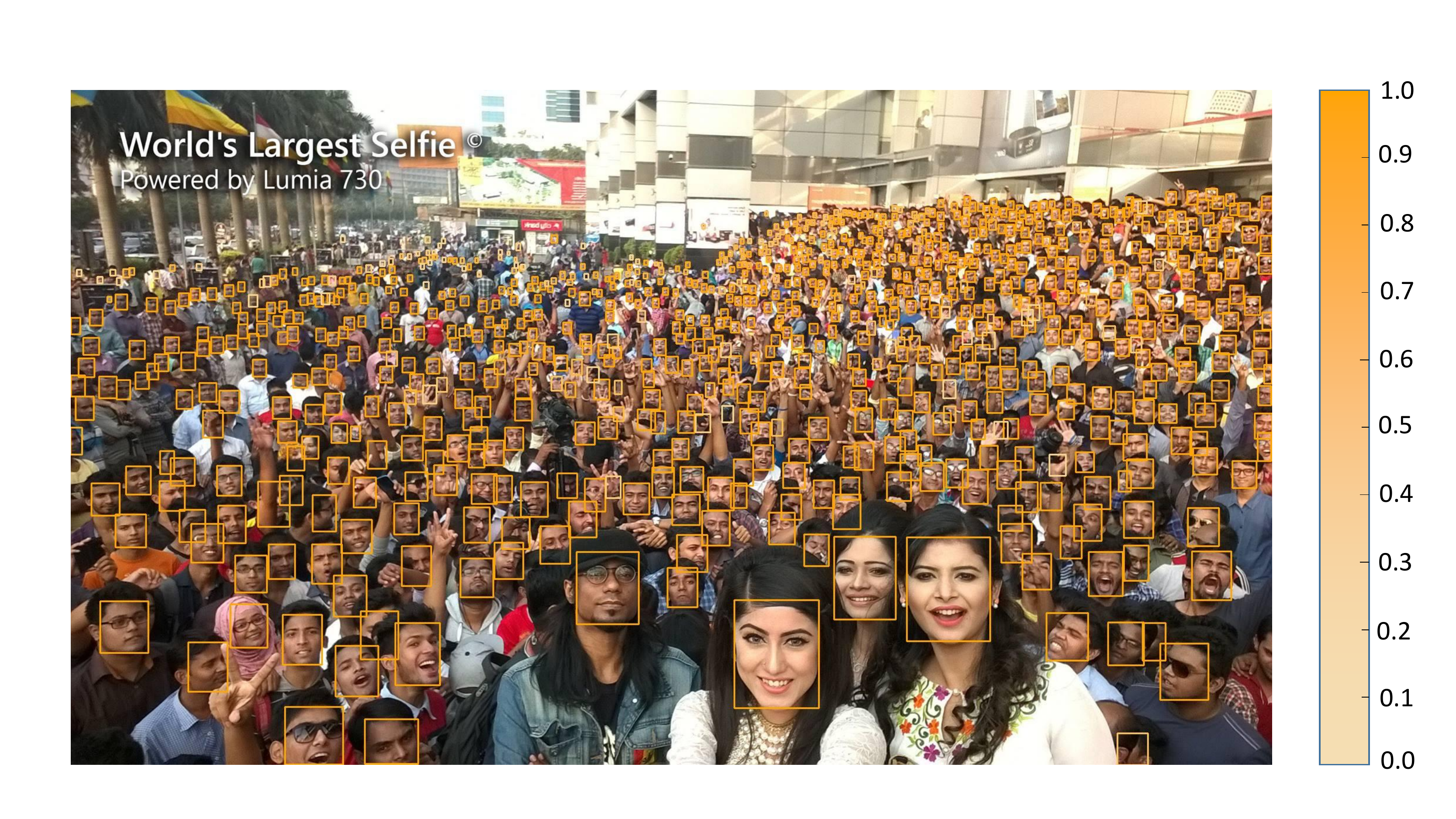}
	\end{center}\vspace{-0.3in}
	\vspace{+0.2in}
	\caption{Example of face detection with the proposed method. In the above image, the proposed method can find 858 faces out of 1000 facial images present. The
		detection confidence scores are also given by the color bar as shown on the right. Best viewed in color.}
	\label{fig:1000face}
\end{figure*}

\begin{figure*}[ht]
	\includegraphics[scale=0.52]{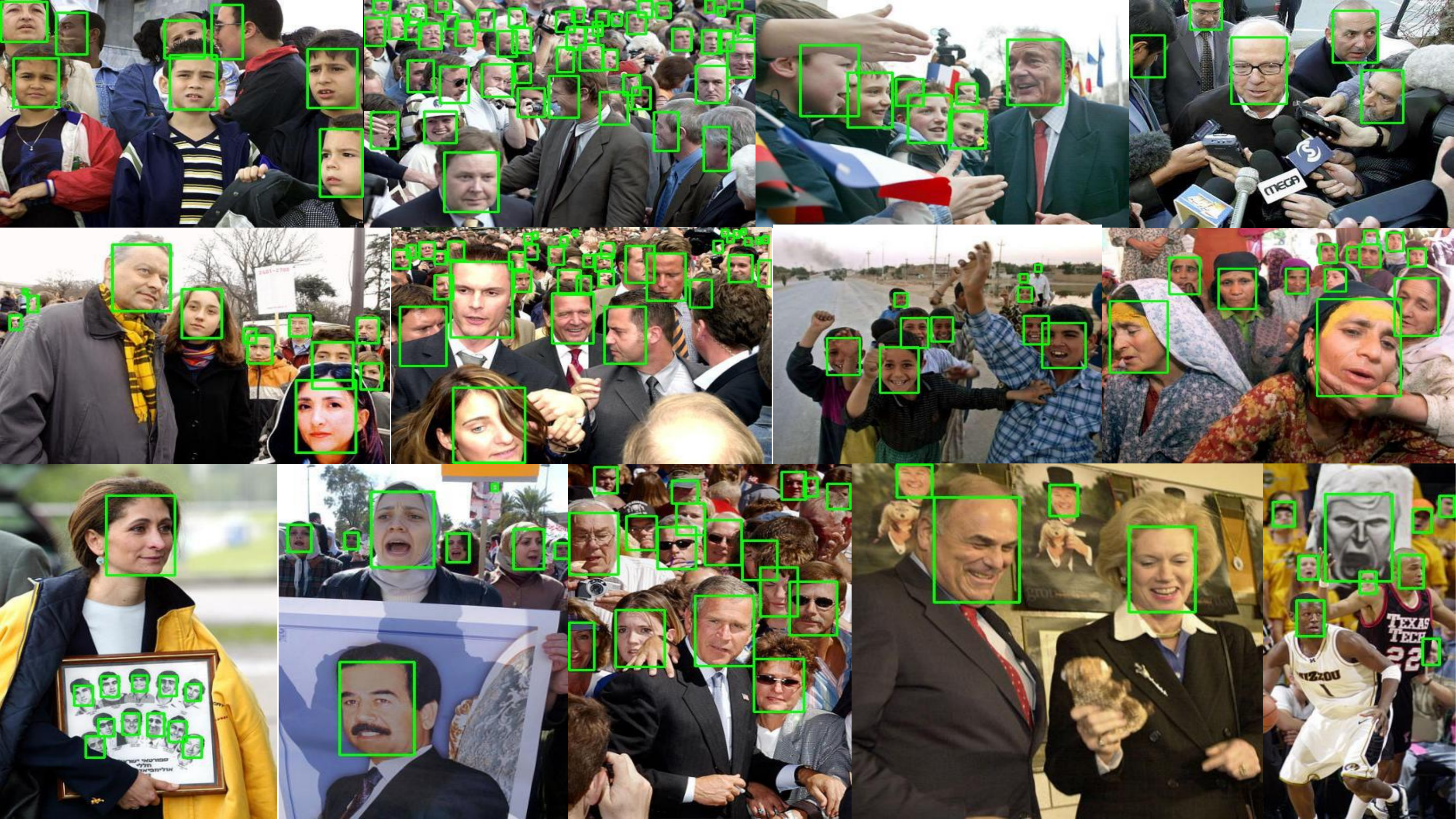}
	\caption{Qualitative results on FDDB. Our model is robust to occlusion and scale variance}
	\label{fig:qr_fddb}
\end{figure*}

\begin{figure*}[ht]
	\includegraphics[scale=0.52, trim={0cm 0cm 0cm 0cm},clip]{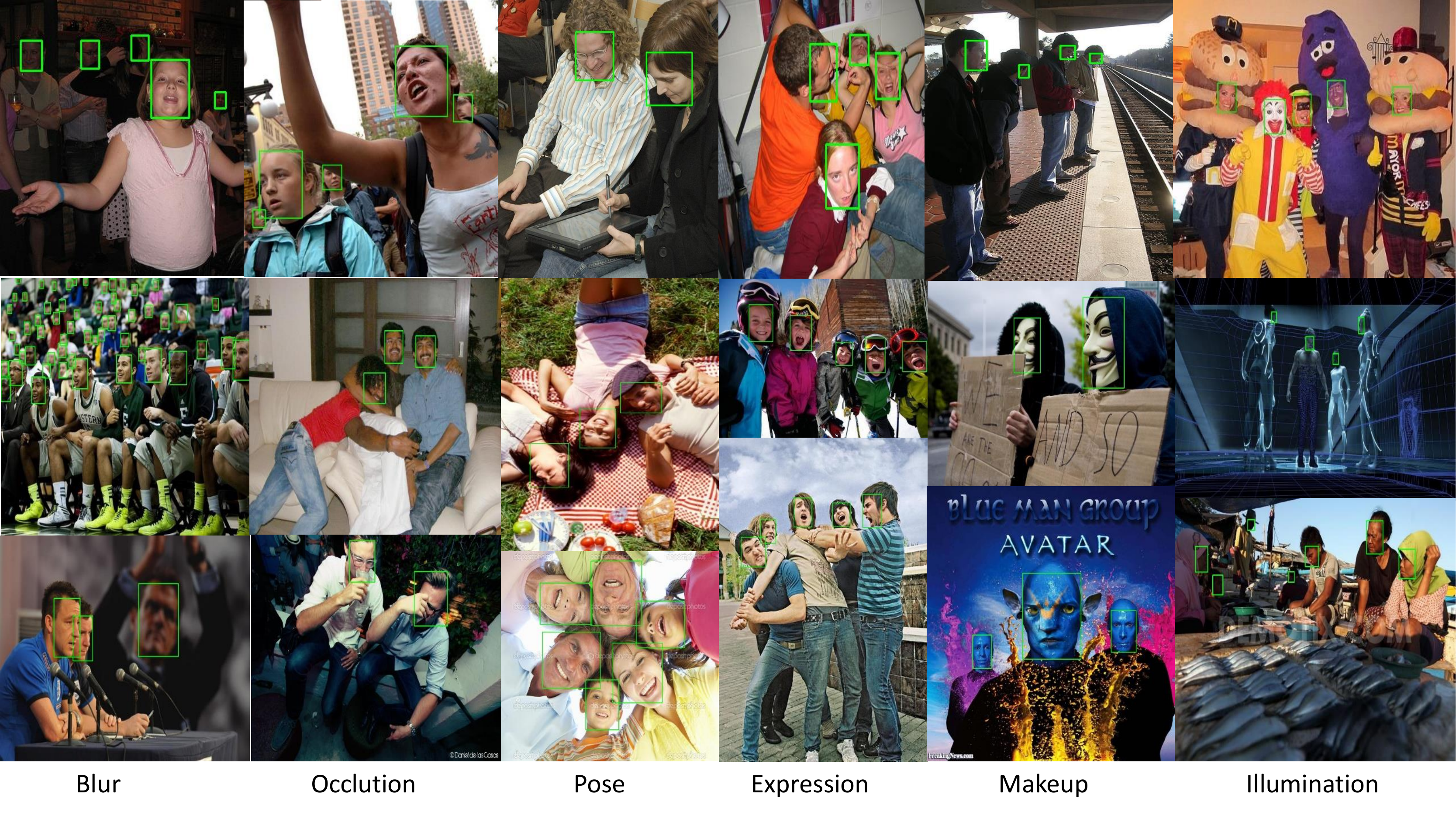}
	\caption{Qualitative results of our FANet. Our model is robust to blur, occlusion, pose, expression, makeup, illumination, etc.}
	\label{fig:qr}
\end{figure*}

In this section, we demonstrate some qualitative results of our FANet,
including World's Largest Selfie, Figure. \ref{fig:1000face}, in which our FANet can find 858 faces out of 1000 facial images present, results on FDDB \cite{fddbTech} dataset, Figure. \ref{fig:qr_fddb} and results under various conditions, Figure. \ref{fig:qr}. With hierarchical loss training scheme, our FANet is very stable, the variance is within 0.2\%.

\end{document}